\documentclass[11pt]{article}
\PassOptionsToPackage{table}{xcolor}

\usepackage[preprint]{acl}

\usepackage{times}
\usepackage{latexsym}

\usepackage[T1]{fontenc}

\usepackage[utf8]{inputenc}

\usepackage{microtype}

\usepackage{inconsolata}

\usepackage{graphicx}
\usepackage{booktabs}
\usepackage{amsfonts}
\usepackage{amsmath}
\usepackage{xcolor}
\usepackage{enumitem}
\usepackage{tabularx}
\usepackage{colortbl}
\usepackage{fontawesome5}   % \faGithub, \faDatabase for the artifact footnote
\usepackage{listings}
\usepackage[colorinlistoftodos,prependcaption,textsize=tiny]{todonotes}

\newcommand\rev[1]{#1}
\newcommand{\framework}{\textsc{KnowSim}}
\newcommand{\dataset}{\textsc{KnowChat}}

\newcommand{\numdataset}{705}

\title{\framework{}: Evaluating Information Calibration in LLM Assistants with User Simulators that Learn}

\author{
  \textbf{Yoonjoo Lee}$^{\heartsuit}$ \hspace{1.2em}
  \textbf{Hyoungwook Jin}$^{\heartsuit}$ \hspace{1.2em}
  \textbf{Tae Soo Kim}$^{\diamondsuit}$ \\[2pt]
  \textbf{Shaoyang Zhang}$^{\heartsuit}$ \hspace{1.2em}
  \textbf{Philippe Laban}$^{\spadesuit}$ \hspace{1.2em}
  \textbf{Q.~Vera Liao}$^{\heartsuit}$ \\[12pt]
  \mdseries
  $^{\heartsuit}$University of Michigan \hspace{1em}
  $^{\diamondsuit}$KAIST \hspace{1em}
  $^{\spadesuit}$Microsoft Research \\[4pt]
  \texttt{\{lyoonjoo\}@umich.edu} 
}
  
\begin{document}

\maketitle

\begin{abstract}
To effectively collaborate with users on knowledge-intensive tasks, Large Language Models (LLMs) must perform \emph{information calibration}: matching content to a user's evolving understanding and cognitive capacity.
Yet user simulators used to evaluate and train LLMs do not explicitly model user knowledge so they neither produce realistic interactions across knowledge levels nor reflect how interactions unfold as that knowledge evolves.
To close this gap, we introduce \framework{}\footnote{\faGlobe\; \href{https://yoonjoolee.com/knowsim/}{\texttt{yoonjoolee.com/knowsim}}}, an evaluation framework built around a user simulator that maintains explicit knowledge states---a graph of Information Units with prerequisite relationships---that evolves under update rules grounded in learning theory. \framework{} computes three metrics (Knowledge Gain, Delivery Calibration, Cognitive Overload) directly from the knowledge state trajectory, reflecting key mechanistic aspects of information calibration. 
We validate \framework{} against \numdataset{} human--AI sessions across two domains, stratified by knowledge level: its rankings align significantly with human judgments (\rev{73--74\%} sign agreement), outperforming three baseline simulators. Applied to 9 LLMs, \framework{} reveals that the best model shifts by user knowledge level---aptitude--treatment interactions invisible to standard evaluation.
% We validate the evaluation framework through a human study spanning two contrasting domains, by collecting \numdataset{} conversation sessions stratified by user knowledge level. \framework{}'s rankings align significantly with human judgments (73--74\% sign agreement), outperforming three baseline simulators on the common metric. We release the dataset and framework.
% We validate \framework{} through a human study, in which participants rated their preference for different assistants. Based on \numdataset{} collected conversations, we confirm that \framework{}'s user simulator aligns significantly with human preference (73--74\% sign agreement), outperforming three baseline simulators. We release the dataset and framework.
\end{abstract}

\vspace{-0.6em}
\begin{center}\small
\faGithub\; \href{https://github.com/yjo2lee/knowsim}{\texttt{yjo2lee/knowsim}} \\[2pt]
\raisebox{-0.15em}{\includegraphics[height=0.95em]{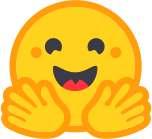}}\; \href{https://huggingface.co/datasets/yjlee36/knowchat-multi-turn-dialogues}{\texttt{yjlee36/knowchat-multi-turn-dialogues}}
\end{center}
\vspace{-0.4em}

\section{Introduction}
\begin{figure*}[t]
\centering
\includegraphics[width=1.00\textwidth]{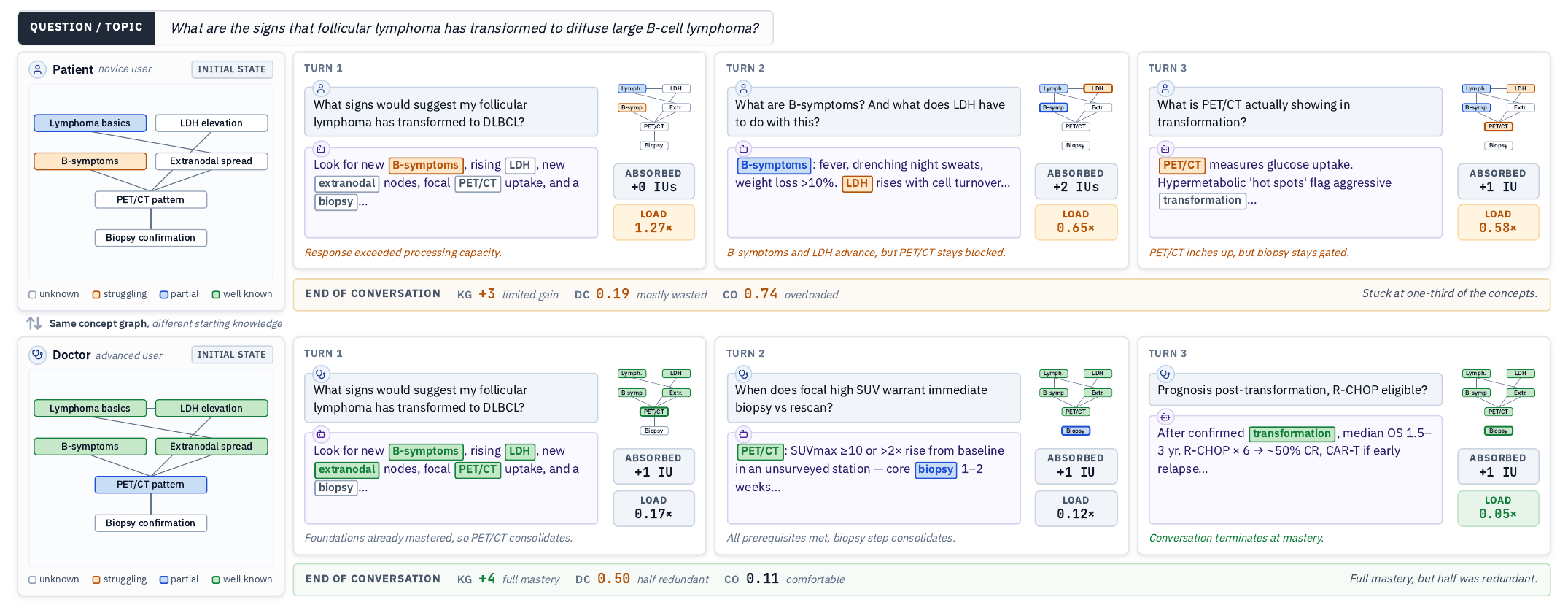}
\caption{\framework{} simulates two users with different knowledge levels asking the same medical question to the same assistant. The novice patient absorbs nothing in Turn~1 (cognitive load exceeds capacity), slowly picks up prerequisites, and ends with limited gain and high overload. The advanced doctor, whose prerequisites are already met, absorbs one concept per turn and reaches full mastery with low overload. Metrics (KG, DC, CO), computed directly from evolving knowledge states, explain \emph{why} the same response helps one user but overwhelms another.}
\label{fig:teaser}
\end{figure*}

As LLMs have grown capable of synthesizing complex information into fluent responses, users increasingly turn to them as collaborative partners in knowledge-intensive tasks---such as understanding medical conditions and analyzing scientific data~\citep{echochamber}. Yet training paradigms such as RLHF~\citep{ouyang2022training} incentivize comprehensive answers in a single message while providing little signal about how information should be sequenced across multiple turns. Consider a doctor and a patient both asking ``What are the signs that follicular lymphoma has transformed to diffuse large B-cell lymphoma?'' 
A doctor can readily understand a dense response that covers symptoms, symptom elevation, and imaging protocols, but a patient who does not know the prerequisite concepts may find it difficult (Fig.~\ref{fig:teaser}).
% A dense response covering symptoms, symptom elevation, and imaging protocols can be readily understood by a doctor, but may be difficult for a patient who has not yet established the prerequisite concepts (Figure~\ref{fig:teaser}). 
Effective collaboration in these tasks requires \emph{information calibration}, not information maximization---delivering the right amount, at the right depth, and in the right sequence for the user's current understanding and cognitive capacity.

Recent work has begun to explore LLM-based user simulators as scalable proxies for human evaluation or feedback in multi-turn settings~\citep{dou-etal-2025-simulatorarena, naous2025flipping, balog2026usersimulationeragenerative, ni-etal-2026-survey}, but current LLM-based simulators are behavioral proxies with two critical limitations. 
First, without knowledge-state modeling, they cannot evaluate information calibration for users at different knowledge levels, only aggregate utility~\citep{chiang2024chatbot}. Second, without tracking how knowledge evolves, they cannot measure objective knowledge gain and instead rely on surface quality of simulated interactions, often via LLM judgments~\citep{zheng2023judging}. This may misleadingly reward models that appear helpful without genuinely supporting how people actually seek information for learning and understanding~\citep{chang2025chatbench}.  

%Without methods to evaluate how user understanding changes in response to information delivery, we cannot distinguish models that genuinely support understanding from those that merely appear useful from a third-party perspective.

% Closing these gaps requires modeling interactions based on knowledge states and how knowledge states evolve with interactions. Similar goals have been tackled in Intelligent Tutoring Systems (ITS): tracking what learners know and how their knowledge evolves to deliver effective scaffolding that targets the knowledge gaps~\citep{corbett1995bkt}. However, knowledge-tracking methods used in ITS depend on hand-crafted (e.g., created by teachers), domain-specific knowledge graphs and predefined curricula to enable knowledge updates, so they cannot be readily transferred to open-ended LLM evaluations~\citep{aleven2006cognitive, matsuda2015teaching}. We draw on ITS principles to develop domain-agnostic, scalable approaches for creating knowledge representations and update rules to power knowledge tracking for LLM evaluation.

We introduce \framework{}, an evaluation framework centered on a knowledge-state-grounded user simulator. The simulator represents the state for user understanding of a given topic as a graph of Information Units (IUs)---self-contained concepts connected by prerequisite relationships. When interacting with an LLM-based assistant, the simulator updates this state by determining how well the assistant explains each IU, whether the user can understand and absorb these IUs based on their current state, and moderates absorption based on cognitive load~\citep{Ausubel2000TheAA}. Evaluation metrics that reflect key aspects of information calibration (Knowledge Gain, Delivery Calibration, Cognitive Overload) are computed directly from the simulator's internal state, providing mechanistic explanations for \textit{why} an assistant helps or hinders a user.

We validate \framework{} through a human study spanning two contrasting domains: math problem solving~\citep{hendrycks2021measuring}, with verifiable reasoning and well-defined prerequisite structure, and expert-level question answering~\citep{Malaviya2023ExpertQAEQ}, where knowledge is more open-ended. We recruit participants stratified by their initial knowledge level (novice / intermediate / advanced) and construct \dataset{}, comprising \numdataset{} conversations in which assistants vary along two axes: information delivery strategy and base model, each paired with subjective ratings and, for MathQA, pre/post knowledge tests. 
\framework{}'s rankings on condition align with human judgments (\rev{73--74\%} sign agreement, $p{=}.003$), with the strongest alignment at the novice level, and outperform three baseline simulators~\citep{dou-etal-2025-simulatorarena}. We then apply \framework{} to benchmark 9 frontier and mid-tier LLMs, revealing that the best model shifts by user knowledge level---tradeoffs invisible to aggregate leaderboards. Our contributions are:
\begin{itemize}[leftmargin=*]
    \item \textbf{Knowledge-state-grounded user simulator.} A simulator grounded in learning theory producing mechanistic metrics (Knowledge Gain, Delivery Calibration, Cognitive Overload) that explain \emph{why} an assistant helps or hinders a user.

    \item \textbf{Human evaluation benchmark with cross-domain validation.} We release \numdataset{} human--assistant sessions across two domains and two comparison axes, with users stratified by knowledge level. Our simulator outperforms three baselines, reaching \rev{73--74\%} sign agreement with human rankings.

    \item \textbf{Frontier model benchmarking.} We apply the validated simulator to 9 LLMs, revealing that the best model shifts by user level---e.g., DeepSeek~V4 maximizes novice knowledge gain while Gemini~3.1~Pro best serves advanced users.
\end{itemize}

\section{Knowledge-Grounded Simulator for Information Calibration Evaluation}

\subsection{Problem Formulation}\label{sec:task}
We study the evaluation of \emph{information calibration} in multi-turn conversations: how well an assistant delivers information matched to a user's existing and evolving knowledge.

\paragraph{Conversation and knowledge state.}
A conversation $C = (u_1, a_1, \ldots, u_T, a_T)$ consists of alternating user messages $u_t$ and assistant responses $a_t$, and focuses on a specific topic determined by a given question $q$ (e.g., math problem, information seeking).
We represent the full understanding of the topic or question as a directed acyclic graph $G_q = (V, E)$ of \emph{Information Units} (IUs): each node $v \in V$ is a self-contained concept relevant to answering $q$, and each edge $(v_i, v_j) \in E$ encodes that $v_i$ is a prerequisite for understanding $v_j$. The user's \emph{knowledge state} at turn $t$ is a labeling $s_t: V \to \{\texttt{unaware}, \texttt{struggling}, \texttt{partial}, \texttt{knows\_well}\}$.  $s$ evolves based on what IUs are provided in assistant turns and the user's ability to absorb or learn them, as described in \S\ref{update}. The initial state $s_0$ is parameterized by the user's knowledge level  $\ell \in \{\text{novice}, \text{intermediate}, \text{advanced}\}$ (Fig. \ref{fig:teaser}).
% which determines the proportion and prerequisite-depth of IUs labeled \texttt{knows\_well} at the outset.

\paragraph{Calibration quality.}
An assistant turn $a_t$ is \emph{well-calibrated} with respect to $s_{t-1}$ if the IUs introduced are (i) \textit{novel:} not already labeled \texttt{knows\_well}, (ii) \textit{understandable:}  prerequisite-reachable given $s_{t-1}$, and (iii) \textit{processable:}  within the user's cognitive absorption capacity at turn $t$. A \emph{well-calibrated assistant} is one whose turns satisfy these conditions consistently as the conversation progresses. Conditions (i)--(iii) are operationalized as the evaluation metrics described in \S\ref{sec:simulator}.

\paragraph{Evaluation task.}
Given an assistant model $M$, a question $q$ with IU graph $G_q$, and a user's current knowledge level $\ell$, we simulate a multi-turn conversation between $M$ and the user initialized at $s_0(\ell)$, and compute calibration metrics from the resulting state trajectory $s_0, s_1, \ldots, s_T$. The metrics are designed to be computable automatically, to remain interpretable in terms of underlying state transitions, and to align with human judgments of calibration quality, which we validate in \S\ref{sec:validation}.

\subsection{User Simulation Method}\label{sec:simulator}
The simulator alternates between user message generation and a state-update pipeline that updates the user's knowledge state based on what knowledge the user can absorb from each assistant response. Full prompts and details are in \S\ref{appendix:user_simulator}.
% The simulator generates user messages and updates the user's knowledge state based on what knowledge the user can absorb from each assistant response. Full prompts, parameter settings, and complete update logic are in \S\ref{appendix:user_simulator}.

\subsubsection{IU Graph Construction}
For each question $q$, the IU graph $G_q$ is grounded in a reference answer $r$: any text that comprehensively covers the concepts relevant to $q$, such as a gold answer for tasks with verifiable solutions or a tutorial or encyclopedia entry for open-ended topics. Nodes in $G_q$ correspond to the concepts required to understand $r$, and edges to prerequisite relations among them. $G_q$ is obtained via LLM-based extraction (\S\ref{appendix:iu_example}) and held fixed across all simulation conditions. To simulate a user at knowledge level $\ell$, we initialize $s_0$ by sampling IUs to be marked \texttt{knows\_well}, \texttt{partial\_understanding}, or \texttt{struggling} according to a ratio $R_\ell$. Sampling follows topological order so that foundational prerequisites are preferentially selected.
% In our experiment (Section~\ref{sec:validation}), when simulating a participant (from the human study) of a given knowledge level, the participant's prescreening knowledge score is mapped to the corresponding $\ell$ and $R_\ell$.

\subsubsection{User Message Generation}
At each turn, the user simulator LLM generates $u_t$ conditioned on the conversation history and the current state $s_{t-1}$. Generation enforces knowledge-consistent behavior: the user cannot spontaneously mention \texttt{unaware} concepts, makes realistic errors on \texttt{struggling} concepts, and applies \texttt{knows\_well} concepts correctly. Based on the \rev{local neighborhood density of} IUs at \rev{\texttt{partial\_understanding}} or above in $s_{t-1}$, we also determine an \emph{articulation mode} (\texttt{explicit} / \texttt{vague} / \texttt{deferential}) that further shapes query specificity---confident users ask precise questions, while less confident users ask broader ones.

\subsubsection{Knowledge State Update per Turn}
\label{update}
After each turn pair $(u_t, a_t)$, the user's knowledge state $s_t$ is updated via: (1) a \emph{signal extraction} phase that reads per-IU interaction signals from $u_t$ and $a_t$ relative to $s_{t-1}$, and (2) a \emph{state update} phase that applies update rules grounded in learning theory.
%After each interaction turn pair $(u_t, a_t)$, the user's knowledge state $s_t$ is updated via a two-phase pipeline.

\paragraph{Signal extraction.}
A single LLM call over $G_q$, $s_{t-1}$, and $(u_t, a_t)$ assigns two labels to every IU $v \in V$. From the assistant turn, $v$'s \emph{teaching quality} is labeled as either \texttt{well\_explained} (e.g., definition derivation, worked examples, etc.), \texttt{shallow} (i.e., only named or partially treated), or \texttt{not\_mentioned}. From the user message, $v$'s \emph{engagement} is labeled as \texttt{reasoning} (i.e., user performs their own reasoning on the IU), \texttt{articulation} (i.e., \rev{user restates content the assistant already produced}), or \texttt{none}.

\paragraph{State update.}
Given the extracted signals, $s_t$ is computed deterministically under two sets of rules: \emph{drivers} of absorption in the assistant provided information, and \emph{constraints} from the user's cognition.

\paragraph{Driver of absorption.} State transitions are driven by the assistant's \emph{teaching quality}. \texttt{well\_explained} IU advances the user's state by one step. A \texttt{shallow} mention advances only the \texttt{unaware}$\to$\texttt{struggling} transition and has no effect at higher states---merely naming a concept creates awareness, but advancing understanding requires substantive explanation \citep{vygotsky1978mind}. \texttt{not\_mentioned} IUs are unchanged.
% A \texttt{shallow} mention advances only the initial transition from \texttt{unaware} to \texttt{struggling}, reflecting that surface exposure creates awareness but not understanding \citep{vygotsky1978mind}. \texttt{not\_mentioned} IUs are unchanged.

\paragraph{Constraints on absorption.} Three constraints bound the state transitions to simulate human cognitive limits. \emph{Prerequisite ceiling} caps each IU's attainable state by the current mastery of its prerequisites: a user cannot truly grasp $v_j$ without first understanding foundational $v_i$ \citep{Ausubel2000TheAA}. \emph{Cognitive overload} computes a state-weighted load $L_t$ over IUs mentioned by both the user and assistants, where less-known IUs contribute higher load and user \texttt{reasoning} attempts also contribute more. When $L_t$ exceeds capacity $\tau$, a sigmoid dropoff continuously reduces the magnitude of upward transitions, and fractional progress lost in any turn carries forward so partial exposure compounds across turns \citep{sweller1994cognitive, sweller2011cognitive}. \emph{Monotonicity} prevents knowledge states from regressing, ensuring that once a concept is learned it is not forgotten within the session. Full definitions and parameter values are in \S\ref{app:update_rules}.

% \paragraph{Constraints on absorption.} The drivers above are bounded by three constraint rules to simulate realistic human cognitive limits. The \emph{prerequisite ceiling} caps an IU's attainable state based on the current mastery of its prerequisites, reflecting the theory that a user cannot truly grasp $v_j$ without first understanding foundational concept $v_i$ \citep{Ausubel2000TheAA}. The \emph{cognitive overload cap} calculates a state-weighted load across all IUs mentioned by both the user and assistant. If the load exceeds a predefined capacity, absorption is suppressed across the affected concepts to model cognitive overload \citep{sweller1994cognitive, sweller2011cognitive}. Finally, \emph{monotonicity} generally prevents knowledge states from regressing, with one exception: if a user's reasoning attempt is confirmed incorrect without an accompanying explanation, the state for that IU degrades to \texttt{struggling}.

% These signals feed five priority-ordered update rules grounded in learning
% theory: \emph{prerequisite ceiling} (Ausubel's Assimilation Theory), 
% \emph{cognitive overload cap} (Cognitive Load Theory), \emph{monotonicity},
% \emph{teaching-quality advance} (Vygotsky's ZPD), and \emph{attempt handling}
% (Testing Effect). Concretely, well-explained concepts advance further than
% shallow mentions, prerequisite-blocked IUs cannot exceed their ceiling, and
% turns whose total cognitive load exceeds capacity yield diminished absorption
% across the affected IUs. Full rule definitions, the load formula, and edge 
% cases are in Appendix~\ref{app:rules}.

\subsubsection{Conversation Termination}\label{sec:termination_rule}
Rather than imposing a fixed turn limit, the simulator terminates under two progress-based conditions grounded in the \emph{learning progress} hypothesis~\citep{oudeyer2016intrinsic, loewenstein1994curiosity} that engagement is sustained while progress is perceived and degrades when it saturates or stalls. Termination triggers when the user (i) achieves \emph{mastery} of the target IUs, or (ii) shows \emph{persistent non-progress or overload}---at least two of cognitive overload, no upward state transitions, and no substantive teaching, sustained over a five-turn window. 
Any conversation that triggers neither is capped at a maximum of 15 turns.
% Detection thresholds and window parameters are specified in \S\ref{appendix:user_simulator}.
Detailed parameters for the detection are in \S\ref{app:termination}.

% Rather than using a fixed turn limit, the simulator terminates conversations under three conditions grounded in the \emph{learning progress} hypothesis~\citep{oudeyer2016intrinsic, loewenstein1994curiosity}---that engagement is sustained while the learner perceives progress, and degrades when progress either saturates or stalls with difficulty. Specifically: \emph{mastery}, when the user reaches \texttt{knows\_well} on all IUs or confirms a correct final answer (progress saturates); \emph{persistent cognitive overload}, when overload recurs in at least 2 of the last 3 turns with no more than one upward state transition after a 3-turn warm-up (progress stalls); and \emph{user-initiated termination}, when the user simulator independently signals that the conversation is no longer productive.

\subsection{Evaluation Metrics}
The framework produces three metrics reflecting the outcome of and requirements for calibration quality computed deterministically from the state trajectory $s_0, \ldots, s_T$ and the per-turn classifications from knowledge state update:
% \begin{itemize}
%     \item \textbf{Normalized Knowledge Gain (NKG)}---total state advancement divided by the maximum 
%     advancement possible from $s_0$, with overload-blocked IUs excluded 
%     from the numerator.
%     \item \textbf{Information Calibration (IC)}---fraction of 
%     assistant-explained IUs that were both new (not yet \texttt{knows\_well})
%     and prerequisite-reachable; penalizes over-explanation and 
%     under-scaffolding simultaneously.
%     \item \textbf{Perceived Overload (PO)}---average per-turn cognitive load
%     exceeding the user's absorption capacity.
%     \item \textbf{Engagement (ENG)}---mean per-turn 
%     $\text{Familiarity} \times \text{Novelty}$.
% \end{itemize}

% \paragraph{Normalized Knowledge Gain (NKG).} 
% Total knowledge state advancement divided by the maximum advancement possible from $s_0$, with overload-blocked IUs excluded from the numerator:
% \begin{equation}
%     \text{NKG} = \frac{\displaystyle\sum_t \sum_{v \in A_t} \max\bigl(0,\, s_t(v) {-} s_{t-1}(v)\bigr)}
%                      {\displaystyle\sum_{v:\, s_0(v) < s_{\max}} \bigl(s_{\max} {-} s_0(v)\bigr)}
% \end{equation}
% where $A_t$ is the set of IUs not cut off by cognitive overload at turn $t$, and states are mapped to ordinals \texttt{unaware}=0 through \texttt{knows\_well}=3.

\paragraph{Knowledge Gain (KG).}
Total ordinal advancement summed over IUs, with states mapped \texttt{unaware}{=}0, \texttt{struggling}{=}1, \texttt{partial}{=}2, \texttt{knows\_well}{=}3:
% The total knowledge state advancement from the initial state to the final state, measured as the sum of per-IU ordinal gains:
\begin{equation}
    \text{KG} = \sum_{v \in V} \max\bigl(0,\, s_T(v) - s_0(v)\bigr)
\end{equation}
% where states are mapped to ordinals \texttt{unaware}{=}0, \texttt{struggling}{=}1, \texttt{partial}{=}2, \texttt{knows\_well}{=}3. NKG captures the raw magnitude of learning progress across all IUs.

\paragraph{Delivery Calibration (DC).}
\rev{How well the assistant delivers teachable content that the user can absorb, measured by the harmonic mean of precision over explained IUs ($E_t$) and recall over teachable IUs.}
\begin{equation}
    \rev{\text{DC} = \frac{2PR}{P+R}}
\end{equation}
\rev{where $P = \sum_t |A_t| / \sum_t |E_t|$, $R = \sum_t |A_t| / \sum_t |Z_t|$, $A_t = \{v \in E_t : s_{t-1}(v) < \texttt{kw} \wedge \text{ceil}_{t-1}(v) \geq \texttt{pu} \wedge s_t(v) > s_{t-1}(v)\}$, and $Z_t$ is the set of IUs teachable at turn $t$,} with \texttt{pu} = \texttt{partial\_understanding} and \texttt{kw} = \texttt{knows\_well}. \rev{DC penalizes redundant, prerequisite-inappropriate, and ultimately unabsorbed delivery.}

\paragraph{Cognitive Overload (CO).} 
Average per-turn load relative to capacity, saturating at one:
% The average per-turn cognitive load relative to the user's absorption capacity (saturating at one), capturing the user's ability to process the conversation:
\begin{equation}
    \text{CO} = \frac{1}{T} \sum_t \min\!\left(1,\; \frac{L_t}{\tau}\right)
\end{equation}
where $L_t$ is a state-weighted load over IUs mentioned and user attempts at turn $t$, and $\tau$ is the capacity threshold (details in \S\ref{appendix:user_simulator}).

% The metrics are designed to be computable automatically, to remain interpretable in terms of underlying state transitions, and to align with human judgments of calibration quality, which we validate in \S\ref{sec:validation}.
We design the metrics to be automatically computable, remain interpretable in terms of underlying state transitions, and align with human judgments of calibration quality, which we validate in \S\ref{sec:validation}.

\section{\dataset{}: A Benchmark of Knowledge-Stratified 
Conversations with Learning Experience Annotations}
\label{sec:data_collection}

We collect conversations between real users and AI assistants to serve as ground truth for validating our simulator. The dataset captures both objective learning outcomes and subjective perceptions of users with varying levels of knowledge interacting with assistants that employ different information-delivery strategies and assistant models.

\subsection{Tasks and Conditions}

We design four study arms crossing two factors---comparison target (\textit{strategy} vs.\ \textit{model}) and task (\textit{math} vs.\ \textit{ExpertQA})---each drawing a separate pool of participants in Prolific~\citep{prolific}. As shown in Table~\ref{tab:study_arms}, the \textit{strategy} arms compare conditions of different delivery strategies by fixing the base model; the \textit{model} arms hold the strategy fixed and vary the conditions of base models.

\paragraph{Tasks.}
For \emph{competition mathematics}, we use 15 problems from the MATH dataset~\citep{hendrycks2021measuring} at difficulty level 5, spanning five topic categories (counting \& probability, geometry, intermediate algebra, number theory, precalculus); within each category, we select three problems requiring distinct IU graphs to eliminate cross-session transfer. 
For \emph{ExpertQA}~\citep{Malaviya2023ExpertQAEQ}, participants select a professional domain they are interested in (e.g., visual arts, education, psychology) and receive relevant questions. We prioritize breadth: each participant encounters three distinct questions, collectively covering 150 questions.

\paragraph{Conditions.}
The \textit{strategy} arms compare three information delivery strategies. \textbf{Adaptive} infers the user's knowledge level from conversation cues and adjusts explanation depth accordingly. \textbf{Comprehensive} delivers complete, well-structured explanations independent of user signals. 
\textbf{Socratic} guides by asking thought-provoking questions unless the user wants confirmation or struggles.
\textit{Model} arms compare \textbf{GPT-5.4}, \textbf{Claude Opus 4.7}, and \textbf{Gemini 3 Pro}, all under the simple strategy without much instruction. Full system prompts are in \S\ref{app:strategy_prompts}.

\begin{table}[t]
\centering
\small
\caption{Study arms overview. Each arm recruits ${\sim}$60 participants (min: 50, max: 72). }
\label{tab:study_arms}
\setlength{\tabcolsep}{4pt} 
\begin{tabularx}{\columnwidth}{@{}cccc@{}}
\toprule
Arm & Task & Fixed & Comparison Target \\
\midrule
Math $\times$ Strat. & MATH (L5) & Model & Strategy \\
Math $\times$ Model & MATH (L5) & Strategy & Model \\
Exp. $\times$ Strat. & ExpertQA & Model & Strategy \\
Exp. $\times$ Model & ExpertQA & Strategy & Model \\
\bottomrule
\end{tabularx}
\end{table}
\subsection{Data Collection Procedure}

\paragraph{Step 1. Recruitment and stratification.} 

Each arm recruits ${\sim}$60 Prolific workers, stratified into novice, intermediate, and advanced knowledge levels to enable comparison of simulator and human outcomes within each level. \rev{For both tasks, stratification uses a 10-item domain-specific prescreening test (design in \S\ref{app:question_creation}).}

\paragraph{Step 2: Assigning conditions.} Each participant completes three sessions, one per condition (three strategies or three models, depending on the arm). \emph{Math} uses a fixed pool of 15 problems so that validated pre/post-tests can be administered for objective knowledge gain. We rotate condition--problem pairings across participants to control for ordering and item effects. \emph{ExpertQA} instead prioritizes domain coverage breadth, so each condition draws from a large pool of problems without repetition. 

\paragraph{Step 3. Per-session protocol.} 
\emph{Math} sessions follow a pre-test $\to$ conversation $\to$ post-test $\to$ subjective ratings flow, while \emph{ExpertQA} sessions omit the pre- and post-tests. In both, participants engage in a free-form conversation with the assigned assistant to learn to solve the problem (Fig.~\ref{fig:interface-math}).

% or \emph{math} sessions, participants first complete a 7-item multiple-choice pre-test to capture their knowledge on the given problem. They were instructed to learn to solve the problem by having a free-form conversation with the assigned assistant (Fig.~\ref{fig:interface-math}). After that, they complete the same 7-item post-test, and a set of subjective ratings (see annotations below). For \emph{ExpertQA sessions}, the protocol is the same except for omitting the pre- and post- tests.
% ---constructing a dedicated knowledge test for each of the  expert-authored questions would not scale as our ExpertQA experiment prioritizes coverage breadth.

\subsection{Annotations and Quality Control}
Each session in \dataset{} includes the full conversation traces, annotated with participants' subjective ratings on (10-point scale): \emph{Delivery Calibration}, \emph{Cognitive Overload}, and \emph{Interaction Quality} (all questions in \S\ref{app:ratings}). \emph{Math} sessions additionally include \emph{Knowledge Gain}, $\text{KG} = (\text{Post} - \text{Pre})/(\text{Max}_{post}- \text{Pre})$. We include only participants who complete all 3 sessions, each with at least 2 user turns.
Participants received \$20/hr. 
% The study was approved by IRB; all participants provided informed consent. Per-task dataset statistics (turn counts, conversation length, total cost) are reported in \S\ref{appendix:basic_stats}.
We collected data with informed consent and IRB approval. Per-task dataset statistics are in \S\ref{appendix:basic_stats}.

\section{Experiment Setup}
\label{sec:validation}
% Exact ratios and the mapping function are in Appendix~\ref{app:init}.
We evaluate \framework{} along (i) \emph{alignment with human outcomes}---whether the simulator reproduces human-derived condition rankings across metrics and knowledge levels, and (ii) \emph{comparison with baselines}---how \framework{}'s correlation with human rankings compares to that of baseline simulators on the metric common to all methods. 

%We compare \framework{} against three baseline simulators of increasing sophistication, using \dataset{} (Section~\ref{sec:data_collection}) as the human ground truth.

\subsection{Simulator Methods}
We compare \framework{} against three baselines of increasing sophistication (all using Gemini-3-Flash; full formal definitions in \S\ref{app:baseline_details}):
\textbf{Zero-shot (ZS)} conditions only on a knowledge-level descriptor and conversation history;
\textbf{ZS-CoT} adds a chain-of-thought reasoning step at each turn~\citep{Wei2022ChainOT};
\textbf{ZS-CoT-Prof} further adds a synthetic user profile and initial knowledge state~\citep{dou-etal-2025-simulatorarena}, but does not update it over conversation.

\paragraph{\framework{} (ours).}
Uses Gemini-3-Flash for user-turn generation and per-turn knowledge-state extraction, and GPT-5.2 for one-time IU graph construction. The method is described in \S\ref{sec:simulator}.
% Beyond the initial state $s_0$, the simulator maintains the IU graph $G_q$ and an evolving state labeling $\{s_t\}_{t=0}^{T}$ updated after each assistant turn through our method (Section~\ref{sec:simulator}): $y_u^t \sim \pi_u(\cdot \mid q, s_{t-1}, G_q, H_{t-1}, z^t)$. We use \texttt{gemini-3-flash-preview} for both user turn generation and per-turn signal extraction for knowledge state updates, and \texttt{gpt-5.2} for one-time IU graph construction from the question and reference answer.

\subsection{Simulation Protocol and Metric Coverage}
% The simulator is a per-condition, per-level prediction function rather than a participant-level subject. 
Each simulator is run on every $(q, \ell, c)$ configuration in \dataset{}, where $q$ is a question, $\ell$ a knowledge level, and $c$ a condition. For \framework{}, the initial knowledge state $s_0(\ell)$ is instantiated by sampling from the IU graph $G_q$ according to level-specific ratios (\S\ref{sec:simulator}). All conversations run to a maximum of 15 turns. For \framework{}, we determine the end turn using the termination rule in  \S\ref{sec:termination_rule}. For baseline simulators, we follow SimulatorArena and use an LLM judge to post-hoc identify the natural ending turn~\citep{dou-etal-2025-simulatorarena}.

We compare human and simulator outcomes at the $(\ell, c)$ \emph{cell-mean} level: for each cell, the human mean aggregates all participants assigned to that $(\ell, c)$ combination, and the simulator mean aggregates all simulated conversations at the same configuration. This tests whether the simulator reproduces the population-level effect of each condition within each knowledge-level stratum.
% Each simulator method is then run on every $(q, \ell, c)$ configuration, where $c$ denotes a condition, so that every question is traversed by all three levels once per condition. Our analyses below compare the simulator outcomes and human outcomes at the matching configurations as in Table~\ref{tab:study_arms}.
% The simulator is a per-condition prediction function rather than a participant-level subject. For each \dataset{} participant $P$ and each condition $c$ that $P$ experienced, we run each simulator method on the same $(item, c)$ pair, with the simulator's initial state derived from $P$'s prescreening score. This yields, for every $(P, c)$ cell, paired human and simulator scores on matched configurations.

Following prior work~\citep{dou-etal-2025-simulatorarena}, all four simulators produce \textbf{Interaction Quality (IQ)} scores via a shared LLM rater $\pi_r$, enabling direct cross-method comparison. \framework{} additionally exposes three \textbf{internal metrics} (KG, DC, CO) computed deterministically from its state trajectory (\S\ref{sec:simulator}); these have no counterpart in baseline simulators. Table~\ref{tab:metric_coverage} summarizes coverage. 

% The four simulators differ in which metrics they can expose. Table~\ref{tab:metric_coverage} summarizes coverage. \framework{}'s \textbf{internal metrics (NKG, DC, CO)} are computed deterministically from its state trajectory and have no counterpart in baseline simulators; the only metric directly comparable across all four simulators is \textbf{Interaction Quality (IQ)}, elicited by an LLM rater $\pi_r$ applied to the simulated conversation~\citep{dou-etal-2025-simulatorarena}. \framework{} produces IQ via the same rater for direct comparability.

\begin{table}[t]
\centering
\small
\caption{Metric coverage. IQ is common across simulators; KG, DC, and CO are unique to \framework{}.}
\label{tab:metric_coverage}
\begin{tabular}{@{}lcccc@{}}
\toprule
Source & KG & DC & CO & IQ \\
\midrule
\framework{} (ours)   & \checkmark & \checkmark & \checkmark & \checkmark  \\
Baselines             & ---        & ---        & ---        & \checkmark \\
Human                 & \checkmark$^*$ & \checkmark & \checkmark & \checkmark  \\
\bottomrule
\end{tabular}
\\[2pt]
\footnotesize{$^*$Human KG from pre/post tests; math sessions only.}
\end{table}

\subsection{Analyses}
We assess alignment using \emph{pairwise sign agreement}: within each $(\text{arm}, \text{knowledge level})$ block, we enumerate all condition contrasts, retain those where the human effect exceeds a non-negligible threshold (Cliff's $\delta \geq 0.15$), and check whether contrasts from the given simulator have the same sign. Counts are tested against chance via a one-sided binomial test. This formulation requires only an ordinal commitment to human direction---appropriate for subjective ratings---and admits exact small-sample inference via a one-sided binomial test ($p = 0.5$) without distributional assumptions. The $\delta \geq 0.15$ threshold excludes near-zero human effects where the ground-truth direction is ambiguous; results are stable across $[0.10, 0.20]$. Following~\citet{dou-etal-2025-simulatorarena}, Spearman $\rho$ correlations between human and simulator cell means corroborate all findings (Appendix~\ref{appendix:results}).
% Following~\citet{dou-etal-2025-simulatorarena}, we also compute Spearman $\rho$ correlations between human and simulator cell means for all analyses, included in Appendix~\ref{appendix:results}, which corroborate all main findings.

On the human side, DC, CO, and IQ are participant self-ratings on a 1--10 scale collected after each session; KG is computed from pre/post knowledge tests (MathQA only). We analyze along two axes: (i)~alignment of \framework{}'s four metrics with human outcomes, reported as metric-aggregated sign agreement per arm (Table~\ref{tab:sign_agreement}) and per (task, level) (Fig.~\ref{fig:per_level_agreement}); and metric specific sign agreement aggregated by arms and levels (Fig.~\ref{fig:per_metric_sign_agreement}); and (ii)~comparison with baseline simulators on IQ---the one metric common to all methods---using the same sign agreement pooled across all arms (\S\ref{sec:valid_baselines}). 
% For the IQ rater $\pi_r$, we select Claude Sonnet 4.6 after comparing three candidate LLMs on human-AI conversations and checking for self-preference bias.
For the IQ rater $\pi_r$, we select Claude Sonnet 4.6 after testing three candidate LLMs and checking for self-preference bias.

\section{Results}\label{sec:results}
We show that (1) \framework{}'s multi-metric signal aligns with human judgment across tasks and knowledge levels (\S\ref{sec:valid_overall}) and (2) it outperforms baseline simulators on the common metric (\S\ref{sec:valid_baselines}).

\subsection{Overall Alignment with Human Judgment}\label{sec:valid_overall}
\begin{table}[t]
\centering
\caption{Sign agreement between \framework{} and human pairwise condition contrasts on high-signal cells ($|\delta_{\text{human}}| \geq 0.15$). Rate indicates the fraction of contrasts where the simulator's direction matches the human Cliff's $\delta$; $p$ is from a one-sided binomial test against chance (50\%).}
\label{tab:sign_agreement}
\small
\begin{tabular}{@{}l ccc@{}}
\toprule
\textbf{Evaluation Setting} & \textbf{Agree / HS} & \textbf{Rate} & $\boldsymbol{p}$ \\
\midrule
MathQA--Strategy   & 13 / 17 & 76\% & .025$^*$ \\
MathQA--Model      & 14 / 20 & 70\% & .058$^\dagger$ \\
ExpertQA--Strategy & 16 / 20 & 80\% & .006$^{**}$ \\
ExpertQA--Model    & 10 / 15 & 67\% & .151 \\
\midrule
MathQA (pooled)    & 27 / 37 & 73\% & .003$^{**}$ \\
ExpertQA (pooled)  & 26 / 35 & 74\% & .003$^{**}$ \\
\bottomrule
\end{tabular}
\end{table}

\textbf{\framework{}'s ranking direction aligns with human judgment across both task domains (Table~\ref{tab:sign_agreement}).}
Pooled across Strategy and Model arms, sign agreement reaches significance on both MathQA (27/37, 73\%, $p{=}.003$) and ExpertQA (\rev{26/35, 74\%, $p{=}.003$}). Per-arm rates range from 67\% to \rev{80\%}, with two of four arms individually significant ($p{=}.025$ and \rev{$p{=}.006$}). At the task level, \framework{} is a directionally reliable proxy, motivating the finer-grained analyses below.\footnote{Significance: one-sided exact binomial vs.\ 50\% chance baseline, with $^\dagger p<.10$, $^* p<.05$, $^{**} p<.01$.}

\begin{figure}[t]
\centering
\includegraphics[width=\columnwidth]{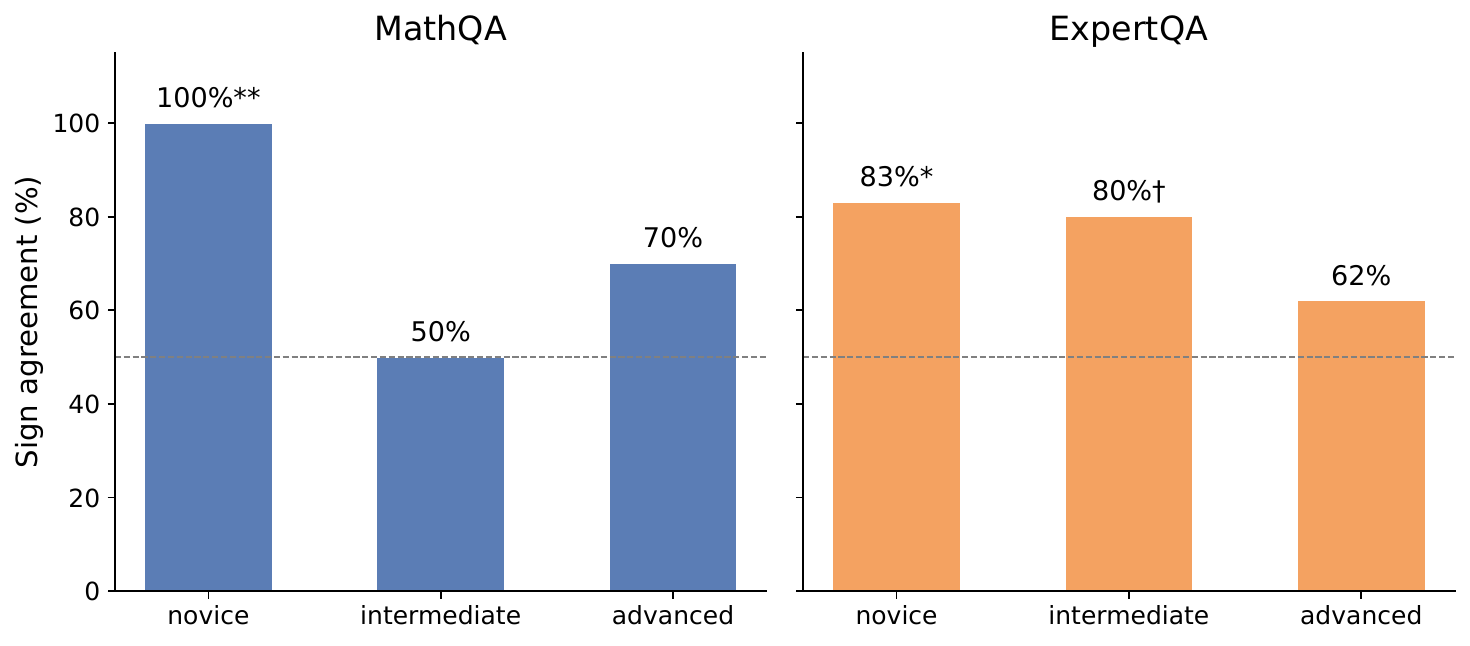}
\caption{Per-level sign agreement (\%) between \framework{} and human rankings, by task (Strategy + Model arms pooled). 
% Dashed line: 50\% chance.
}
\label{fig:per_level_agreement}
\end{figure}

\textbf{Novice level has the strongest alignment in both tasks (Fig.~\ref{fig:per_level_agreement}).}
Agreement at the novice level is 100\% (13/13, $p{<}.001$) on MathQA and \rev{83\% (10/12, $p{=}.019$)} on ExpertQA. Intermediate and advanced levels are more variable: \rev{MathQA intermediate drops to chance (50\%), while on ExpertQA intermediate remains high (80\%, $p{=}.055$) and advanced falls to 62\%.}
We interpret this pattern through the lens of what drives user preferences at each level. 
% [rev] Removed the per-level Spearman citations (combined rho=0.67; advanced
% rho=-0.03/-0.27/-0.08; IQ rho=0.52, p=.085) -- n is 12 per metric (6 for KG),
% too small to carry the claim. Also removed the parenthetical
% "(e.g., intellectual engagement, conversational efficiency)", which those
% correlations were the only support for.
For novice users, comprehension and learning progress---precisely what \framework{}'s knowledge-state model tracks---are likely to dominate session preferences, \rev{and \framework{} reproduces their rankings most reliably on both tasks. At higher knowledge levels, agreement is both lower and less consistent across the two tasks, suggesting} that more knowledgeable users evaluate sessions along dimensions beyond knowledge acquisition that our state-based metrics do not yet model.
%, whereas novices' preferences are well-explained by the learning dynamics \framework{} explicitly captures.
% This cross-task consistency at the novice level suggests \framework{} most reliably captures how lower-knowledge users discriminate among assistants---where KS-aware modeling provides the largest informational margin.

\begin{figure}[t]
\centering
\includegraphics[width=\columnwidth]{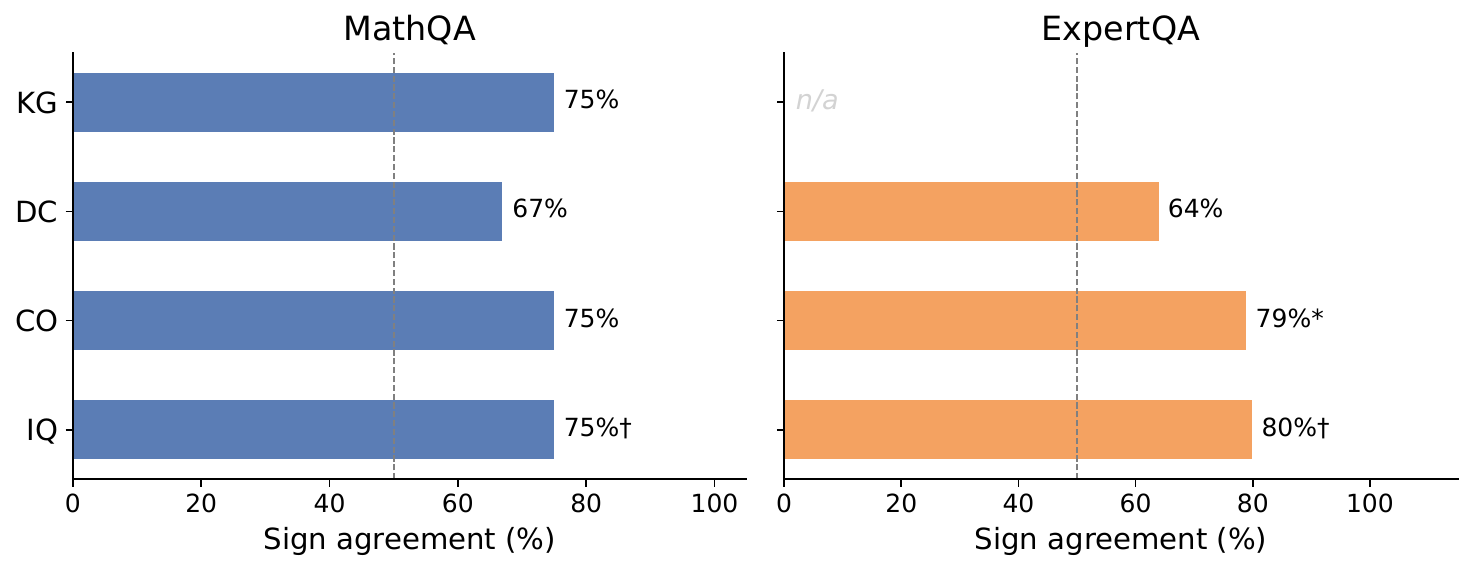}
\caption{Per-metric sign agreement (\%) between \framework{} and human rankings, by task (Strategy + Model arms pooled). KG is not measured on ExpertQA.}
\label{fig:per_metric_sign_agreement}
\end{figure}

\textbf{IQ and CO emerge as the most consistent diagnostic metrics across both tasks (Fig.~\ref{fig:per_metric_sign_agreement}).}
On MathQA, IQ, KG, and CO each reach 75\% agreement (IQ: $p{=}.073$); DC trails at 67\%. On ExpertQA, \rev{IQ (80\%, $p{=}.055$) and CO (79\%, $p{=}.029$) lead}; DC remains the lowest (\rev{64\%}).
Overall, alignment is consistently high across IQ, KG (MathQA), and CO in both tasks, while DC shows a modest gap that may reflect the difficulty of capturing perceived calibration quality through a single subjective rating.
% On MathQA, IQ reaches marginal significance (9/12, 75\%, $p{=}.073$), with NKG and CO close at 75\% (6/8 each); DC trails (6/9, 67\%). On ExpertQA, CO leads (11/14, 79\%, $p{=}.029$) and IQ is close behind (8/10, 80\%, $p{=}.055$); DC remains weakest (7/11, 64\%). The convergence of IQ and CO as the top-performing metrics across both procedural and open-ended tasks suggests these dimensions capture task-general aspects of assistant--user fit, while DC's consistent weakness flags it as the least informative dimension in our metric suite.

\subsection{Comparison with Baselines}\label{sec:valid_baselines}
\begin{figure}[t]
\centering
\includegraphics[width=0.8\columnwidth]{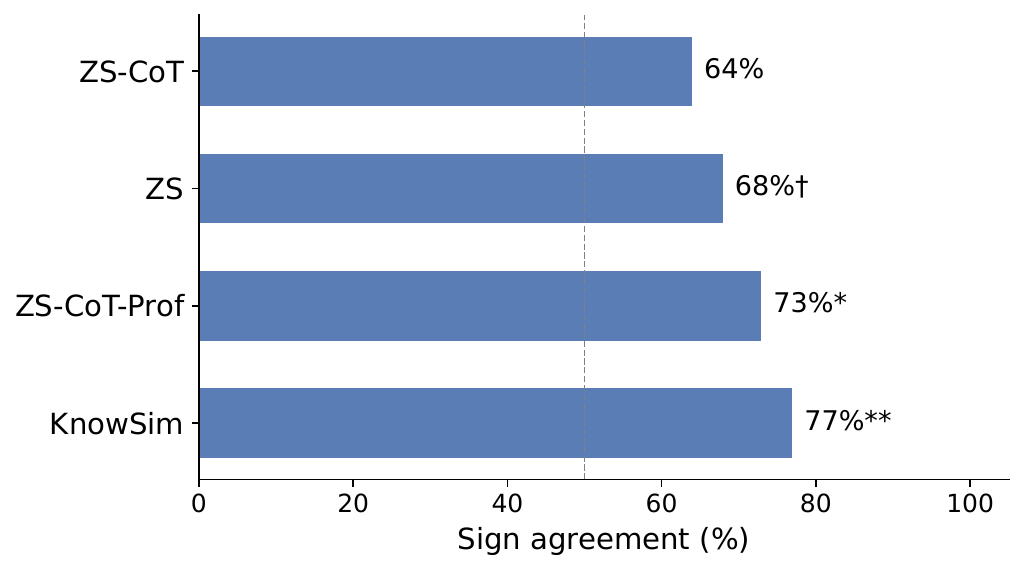}
\caption{Cross-task IQ sign agreement (\%) between each simulator and human rankings, pooled across both tasks and all four arms (\rev{22} signal pairs).
% Dashed line: 50\% chance.
}
\label{fig:baseline_iq_crosstask}
\end{figure}

\textbf{\rev{\framework{} achieves the highest IQ alignment among all simulators} (Fig.~\ref{fig:baseline_iq_crosstask}).} IQ is the only metric that the baselines can compute. Pooling all four arms, \framework{} agrees with human pairwise IQ rankings \rev{77\% of the time ($p{=}.008$), ahead of ZS-CoT-Prof (73\%, $p{=}.026$), ZS (68\%, $p{=}.067$), and ZS-CoT (64\%, $p{=}.143$). Broken down by arm, \framework{} leads or ties ZS-CoT-Prof---the strongest baseline---in all four (Appendix~\ref{app:per_arm_iq}).}

Beyond metric-level agreement, \rev{on MathQA,} \framework{}'s simulated users \rev{show greater overlap with real users in the IUs they engage with than any baseline, as measured by full-turn Jaccard similarity} (Appendix~\ref{app:concept_coverage}).

%Since baselines can only be compared on IQ, they cannot express the per-metric diagnostic signal (CO, DC, NKG; \S\ref{sec:valid_overall}) where \framework{}'s multi-dimensional advantage resides.
\section{Comparing Assistants with \framework{}}
\label{sec:assistant_benchmark}
\definecolor{rankblue}{RGB}{24,95,165}
\newcommand{\rb}[2]{\cellcolor{rankblue!#1!white}#2}
\newcommand{\rbw}[2]{\cellcolor{rankblue!#1!white}\textcolor{white}{#2}}

\begin{table*}[t]
\centering
\caption{Aptitude--treatment interaction patterns: per-level metrics across 9 LLMs. Each cell shows the metric value; $\bar{R}$ is the mean rank across the four metrics within that level. CO$\downarrow$: lower is better. Best value per column is \textbf{bolded}.}
\label{tab:benchmarking-per-level}
\resizebox{0.95\textwidth}{!}{%
\begin{tabular}{l ccccc ccccc ccccc}
\toprule
& \multicolumn{5}{c}{\textbf{Novice}} & \multicolumn{5}{c}{\textbf{Intermediate}} & \multicolumn{5}{c}{\textbf{Advanced}} \\
\cmidrule(lr){2-6} \cmidrule(lr){7-11} \cmidrule(lr){12-16}
\textbf{Model} & KG & CO$\downarrow$ & DC & IQ & $\bar{R}$ & KG & CO$\downarrow$ & DC & IQ & $\bar{R}$ & KG & CO$\downarrow$ & DC & IQ & $\bar{R}$ \\
\midrule
Gemini 3.1 Pro     & 13.2          & \textbf{0.875} & \rev{0.086}          & \textbf{8.39} & \rbw{69}{\textbf{3.25}} & 5.4          & \textbf{0.750} & \rev{0.088}          & 8.35 & \rbw{48}{4.75}          & \textbf{0.7} & 0.873          & \rev{\textbf{0.122}} & 7.96 & \rbw{90}{\textbf{1.75}} \\
Claude Opus 4.7    & 14.3          & 0.882          & \rev{0.090}          & 7.80 & \rbw{69}{\textbf{3.25}} & 5.6          & 0.807          & \rev{\textbf{0.128}} & \textbf{8.39} & \rbw{79}{\textbf{2.50}} & 0.6          & 0.879          & \rev{0.107}          & \textbf{8.30} & \rbw{72}{3.00} \\
DeepSeek V4        & \textbf{15.5} & 0.945          & \rev{0.092}          & 6.50 & \rbw{58}{4.00}          & \textbf{5.7} & 0.835          & \rev{0.121}          & 7.53 & \rbw{55}{4.25}          & 0.5          & 0.890          & \rev{0.109}          & 7.74 & \rbw{55}{4.25} \\
Gemini 3.1 Flash   & 15.1          & 0.936          & \rev{\textbf{0.093}} & 6.83 & \rbw{69}{\textbf{3.25}} & 5.6          & 0.838          & \rev{0.116}          & 7.73 & \rb{34}{5.75}           & 0.5          & 0.912          & \rev{0.109}          & 7.64 & \rb{30}{\rev{6.00}} \\
Claude Sonnet 4.6  & 12.9          & 0.877          & \rev{0.079}          & 7.06 & \rbw{48}{4.75}          & 5.4          & 0.791          & \rev{0.098}          & 7.87 & \rbw{44}{5.00}          & 0.6          & \textbf{0.866} & \rev{0.100}          & 8.00 & \rbw{65}{3.50} \\
GPT-5.4            & 14.9          & 0.940          & \rev{0.088}          & 5.24 & \rb{37}{5.50}           & 5.6          & 0.792          & \rev{0.119}          & 6.86 & \rb{37}{\rev{5.50}}           & 0.4          & 0.903          & \rev{0.102}          & 7.05 & \rb{6}{7.75} \\
Qwen-3.6-35B       & 14.4          & 0.934          & \rev{0.082}          & 5.68 & \rb{41}{5.25}           & \textbf{5.7} & 0.829          & \rev{\textbf{0.128}} & 6.89 & \rbw{51}{4.50}          & 0.5          & 0.892          & \rev{0.081}          & 7.39 & \rb{16}{7.00} \\
GPT-5.4 mini       & 12.6          & 0.946          & \rev{0.075}          & 5.41 & \rb{2}{8.00}            & 5.4          & 0.809          & \rev{0.121}          & 7.13 & \rb{30}{\rev{6.00}}           & 0.5          & 0.901          & \rev{0.114}          & 7.24 & \rb{34}{5.75} \\
Llama-4-Maverick   & 11.8          & 0.927          & \rev{0.071}          & 4.58 & \rb{6}{7.75}            & 5.4          & 0.767          & \rev{0.105}          & 5.64 & \rb{20}{6.75}           & 0.3          & 0.882          & \rev{0.117}          & 6.07 & \rb{30}{6.00} \\
\bottomrule
\end{tabular}%
}
\end{table*}

An important application of user simulators is evaluating how well assistant LLMs calibrate to user populations with different knowledge backgrounds. 
Having established that \framework{} reliably captures information calibration quality (\S\ref{sec:results}), we apply it to benchmark 9 open- and closed-source LLMs. Each LLM is evaluated on 30 MathQA and 30 ExpertQA items, with simulated users at three knowledge levels per item. We report \framework{}'s internal metrics (KG, DC, CO) and IQ in \S\ref{appendix:assistant_benchamrking}.

\paragraph{Overall ranking.}
Table~\ref{tab:benchmarking-overall} (Appendix) reports the level-marginalized ranking across 9 models. \rev{Claude Opus 4.7 leads overall ($\bar{R}$= 2.75), with the strongest DC (0.108) and 2nd-highest IQ (8.16). Gemini 3.1 Pro ranks 2nd ($\bar{R}$=3.50) with the highest IQ (8.24) and lowest CO (0.833)}. DeepSeek V4 ranks \rev{1st} on KG but 8th on CO (0.890), revealing a ``knowledge-dumping'' pattern where aggressive explanation maximizes knowledge throughput while overwhelming the learner. Gemini 3.1 Flash---the mid-tier counterpart to Pro---places 4th overall with strong KG (7.1, rank \rev{2}) but the highest CO (0.895), sharing the overload pattern of high-throughput models. 
% The GPT-5.4 family shows competitive KG but underperforms on IQ (ranks 7--8), suggesting effective knowledge delivery without strong adaptation to user signals.

% Table~\ref{tab:benchmarking-overall} reports the level-marginalized rank of each assistant on each metric, with mean rank across metrics as a synthesis. 
% % Claude Opus 4.7 leads overall (mean rank 2.50), driven by the highest IQ and strong KG, while Claude Sonnet 4.6 achieves the best cognitive-overload control (lowest PO) at the cost of lower knowledge throughput. Notably, the IQ ranking diverges from calibration metrics: DeepSeek V4 ranks first on KG but last on PO, revealing a ``knowledge-dumping'' failure mode where aggressive explanation maximizes surface-level progress while overwhelming the learner.
% Claude Opus 4.7 leads overall (mean rank 2.50), driven by strong DC (rank 3) and the highest IQ (8.16, rank 1). DeepSeek V4 ranks 1st on both KG and 2nd on DC but last on PO (0.843, rank 8), showing ``knowledge-dumping'' in which aggressive explanation maximizes knowledge throughput while overwhelming the user. Gemini 3.1 Pro achieves the best cognitive-overload control (lowest PO=0.811, rank 1) and 2nd-highest IQ (7.85), but delivers less knowledge overall (KG rank 8, DC rank 8). Llama-4-Maverick ranks 2nd on PO (0.817) but last on IQ (5.43, rank 8), suggesting low cognitive load at the cost of shallow interactions. The GPT-5.4 family underperforms on IQ (ranks 5--6) despite competitive KG, suggesting that they effectively advance user understanding without adapting well to user signals.

\paragraph{Aptitude--treatment patterns across knowledge levels.}

% Table~\ref{tab:benchmarking-per-level} breaks down metrics by knowledge level, revealing that \emph{no single model serves all users equally well}. The top-ranked model shifts across levels: DeepSeek V4 dominates novice KG (13.3) and DC (0.131) but at high overload cost (CO=0.937), whereas Gemini 3.1 Pro excels for advanced users (lowest CO=0.721, highest IQ=8.38 at intermediate) yet underserves novices (KG rank 7). Claude Opus 4.7 maintains the highest IQ at intermediate (8.39) and advanced (8.30) but is not the top knowledge-delivery model for novices.

% These rank reversals expose an equity gap invisible to level-agnostic leaderboards: Gemini 3.1 Pro---the best-calibrated model for knowledgeable users---ranks near the bottom on novice knowledge gain, while DeepSeek V4 delivers the most knowledge to novices but overwhelms them. \framework{}'s per-level decomposition makes these tradeoffs explicit, enabling practitioners to select assistants matched to their target user population.

Table~\ref{tab:benchmarking-per-level} breaks down metrics by knowledge level, revealing that \emph{no single model serves all users equally well}. 
The best model for a given metric shifts across levels. For novices, DeepSeek V4 delivers the most knowledge (KG=15.5) but at high overload cost (CO=0.945, second-worst), while Gemini 3.1 Pro achieves the lowest novice CO (0.875) with the highest novice IQ (8.39).
For advanced users the picture inverts: Gemini 3.1 Pro is best-calibrated overall (best $\bar{R}$=1.75) with the highest KG (0.7) and near-lowest overload (CO=0.873), while Claude Opus 4.7 attains the highest IQ at both the intermediate (8.39) and advanced (8.30) levels but is not the top knowledge-delivery model for novices.

These rank reversals expose an equity gap invisible to level-agnostic leaderboards: Gemini 3.1 Pro---the best-calibrated model for advanced users---ranks 6th on novice KG, while DeepSeek V4 delivers the most knowledge to novices but overwhelms them (CO=0.945). \framework{}'s per-level decomposition makes these tradeoffs explicit, enabling practitioners to select assistants matched to their target user population.

Beyond ranking, IU-level tracking lets us decompose why calibration fails: pooled across assistants, novice sessions are dominated by over-reaching (65\% of explained IUs are prerequisite-blocked or overload-wasted), while advanced sessions are dominated by redundant re-explanation (89\%). This directional failure pattern is detailed in \S\ref{app:failure_modes}.

\section{Related Work}
\label{sec:related_work}
% Verbosity ≠Veracity: Demystify Verbosity Compensation Behavior of Large Language Models
% Do Chatbot LLMs Talk Too Much?The YapBench Benchmark

\paragraph{Multi-turn LLM evaluation.}
% Comparing assistants in multi-turn conversations is challenging as each user turn depends on the assistant's prior response, so the user-side input itself becomes a variable that shapes the verdict~\citep{mehri2020unsupervised}. 
Comparing assistants in multi-turn conversations is challenging as the assistant's prior response shapes each user turn, which affects the verdict in return~\citep{mehri2020unsupervised}.
\emph{Static benchmarks}~\citep{zheng2023judging, kwan2024mteval, bai2024mtbench101} and \emph{real-user studies}~\citep{collins2024, ibrahim2024} sit on opposite sides of a fidelity-scalability tradeoff: pre-scripted users cannot react to assistant replies, while live participants do not scale across many assistant-user combinations. Bridging this tradeoff requires user models that interact dynamically with assistants but remain cheap to deploy---motivating our LLM-based simulator paradigm.

% 1

\paragraph{LLM-based user simulators.}

\rev{Beyond prompting, user simulators have been built by augmenting generation with external knowledge~\citep{dhole-2024-kaucus}, self-play and fine-tuning on dialogue data~\citep{kong-etal-2024-platolm, ulmer-etal-2024-bootstrapping}, reinforcement learning of dedicated user language models~\citep{naous2025flipping}, inferring implicit profiles from past interactions~\citep{wang-etal-2025-know}, and agent architectures for search behavior~\citep{zhang2024usimagent}.}
User simulators are often used as \emph{evaluation infrastructure} for specific assistant capabilities~\citep{ibrahim2025, li2024mediq}, such as tool use~\citep{yao2024taubench} and instruction following~\citep{laban2025lostinconv}.
A recent thread adds richer user dynamics: DiscoverLLM~\citep{kim2026discoverllm} models intent formation through a hierarchical tree; HumanLM~\citep{humanlm} aligns latent psychological states (e.g., belief, emotion) with ground-truth responses via RL; and CollabLLM~\citep{wu2025collabllm} forward-samples simulated trajectories to compute multi-turn-aware rewards for model training. 
We extend this research thread by modeling users' evolving \emph{knowledge}, enabling novel metrics for information calibration.

\paragraph{Student simulation and knowledge state modeling.}

A growing body of work uses LLMs to simulate students for educational AI research~\citep{llm-student-survey}: simulating cognitive levels and learning dynamics~\citep{embracing-imperfection-acl, learning-dynamics-2025}, generating student errors~\citep{jin2024teach, ross2025learning}, and auditing simulator quality with human teachers~\citep{can-llms-simulate-learners-bea} or automated metrics~\citep{scarlatos2026simulated}. \rev{However, these efforts typically characterize learners by static knowledge profiles rather than explicitly tracking how their knowledge changes over an interaction.}
% We address the limitation by adopting (1) knowledge tracing~\citep{corbett1995bkt, piech2015dkt} that models students' mastery as latent states updated by observed responses, and (2) intelligent tutoring systems~\citep{anderson1995cognitive-tutor, vanlehn2011} that scaffold instruction through expert-curated prerequisite graphs. We extend the two educational AI concepts to simulate users in open-ended LLM dialogues for assistant evaluation.
\rev{Modeling knowledge as an evolving state has a long history in educational AI. Student modeling represents mastery over structured prerequisite relations~\citep{anderson1995cognitive-tutor, vanlehn2011}, while knowledge tracing updates estimates of mastery based on learners' engagement with material~\citep{corbett1995bkt, piech2015dkt}, with recent work extending this idea to open-ended tutor--student dialogue~\citep{scarlatos2025dialoguekt}. Dialogue systems similarly maintain latent user states across turns, for example through belief trackers that estimate distributions over task-relevant slot values~\citep{mrksic-etal-2017-neural}. We build on these traditions to model evolving knowledge in open-ended LLM dialogue. Unlike dialogue-based knowledge tracing, which infers mastery from observed tutor--student exchanges, our simulator updates the user's knowledge from information received during interaction and uses that state to generate subsequent behavior. IU graphs represent conceptual understanding and prerequisite relations among concepts.}

\section{Conclusion}
We presented \framework{}, a knowledge-state-aware user simulator that models how understanding evolves across multi-turn conversations via an internal knowledge representation, and \dataset{}, a benchmark of \numdataset{} knowledge-stratified human--AI conversations across two task domains with learning outcome annotations. \framework{} aligns with human judgment at \rev{73--74\%} sign agreement and outperforms baseline simulators. Across 9 frontier LLMs, it reveals aptitude--treatment interactions invisible to aggregate leaderboards, where the best model shifts by user knowledge level.

% Benchmarking with \framework{} reveals information miscalibration in SOTA assistants for users with different knowledge levels, which existing leaderboards miss.

\section*{Limitations}
Our simulator models information-processing constraints---cognitive load, prerequisite-gated absorption, and overload-driven termination---but underspecifies motivational and affective dynamics such as frustration and boredom; this manifests in KG saturation for advanced simulated users. 
The simulator nonetheless captures the ATI crossover in Top-1 strategy ranking across Novice, Intermediate, and Advanced learners, and integrating motivational dynamics drawn from expectancy--value or flow theory remains future work.

A second scope limitation is that our evaluation covers two knowledge-intensive domains (math problem-solving and ExpertQA); extending the framework to domains with more open-ended success criteria, such as creative writing or exploratory data analysis, is ongoing work. 
 
While our simulator can express within-group variation by tuning the KS ratio per user, estimating an individual's initial KS state from observable behavior remains challenging in open-ended conversational settings, despite progress in knowledge tracing for structured tutoring contexts~\citep{corbett1995bkt, piech2015dkt}. Bridging this gap with our group-level cognitive abstractions is a promising future research direction.

On the implementation side, our simulator requires multiple LLM calls per turn for IU absorption analysis and user response generation, which constrains evaluation throughput when scaling to many assistant models; since the structured components (IU graph, prerequisite gating, load blend) are deterministic and inexpensive, distilling only the LLM-based modules into smaller fine-tuned simulators could substantially reduce cost without sacrificing alignment. 

Finally, having validated our simulator as an evaluation proxy that produces human-aligned assistant rankings, applying it as a training signal---for instance, as a reward model in RL fine-tuning of assistants \citep{wu2025collabllm, kim2026discoverllm}---is a natural extension we leave for future work.

\section*{Ethical Considerations}
\textbf{Risks and societal impact.} \framework{} is an evaluation proxy and should not be used as a substitute for human studies. Over-reliance on simulator-derived rankings can lead practitioners to deploy assistants that score well in simulations but may underserve real users, particularly at knowledge levels where our metrics may align less with human judgment. Our own results surface an equity dimension: models that maximize novice knowledge gain may do so at high cognitive overload, and the best-calibrated model for advanced users underserves novices. Selecting assistants on aggregate scores alone could therefore systematically disadvantage less-knowledgeable users.

\textbf{Artifacts and licensing.} We release \dataset{} (\numdataset{} human--assistant sessions) and the simulator code for non-commercial use under CC BY-NC 4.0. We build on the MATH dataset~\citep{hendrycks2021measuring} and ExpertQA~\citep{Malaviya2023ExpertQAEQ}. Both are publicly released for research, and our use is consistent with their intended research use. Model outputs were obtained via each provider's API under their respective terms of service.

\textbf{Privacy and content.} Human-assistant sessions were collected via Prolific, which exposes only anonymous participant IDs and these are not retained in the final collected data. We collected no names or other personally identifying information. We manually reviewed a sample of sessions and found no personally identifying or offensive content.

% \section*{Limitations}
% TODO: Add limitations section

% \section*{Acknowledgments}
% TODO: Add acknowledgments

\bibliography{references}
\clearpage
\appendix

\section{User Simulator Details}\label{appendix:user_simulator}

This appendix supplies the prompts, parameter values, and worked-example backing the user simulator described in \S\ref{sec:simulator}. The subsections mirror the order in \S2: IU graph construction (\S\ref{app:iu_extraction}), initial knowledge state (\S\ref{app:init}), user message generation (\S\ref{app:user_prompt}), signal extraction (\S\ref{app:phase_b_prompt}), state update rules (\S\ref{app:update_rules}), conversation termination (\S\ref{app:termination}), and a worked IU graph example (\S\ref{appendix:iu_example}).

\subsection{IU Graph Construction}\label{app:iu_extraction}

We use \texttt{gpt-5.2} for the one-time extraction call per $(q, r)$. The math and ExpertQA prompts share structure with different examples or details. We present a unified prompt below with annotations \texttt{[M]} for Math-only content, \texttt{[E]} for ExpertQA-only content, and labeled worked examples per domain.

\noindent
\textbf{IU Graph Extraction Prompt (unified Math + ExpertQA)}
\begin{lstlisting}[escapechar=@]
You are an expert at analyzing complex questions and their answers to extract structured knowledge representations. Given a question and its reference answer, extract an Information Unit (IU) Graph - a directed acyclic graph that represents all the pieces of understanding a person needs to fully comprehend the answer.

[E] ## Domain Context
[E] Field: {field}
[E] Specific field: {specific_field}
[E] Use this context to calibrate terminology and concept granularity. Do not restrict the graph to concepts from this field only - include any cross-domain prerequisites that are genuinely necessary for comprehension.

## What is an Information Unit (IU)?

An Information Unit is a self-contained piece of understanding that:
1. Independently assessable - you can determine whether someone "gets it" as a standalone unit
2. Explainable in 2-4 sentences - not a single fact, not an entire topic
3. Has prerequisite relationships with other IUs - understanding some IUs requires first understanding others

[M] IUs can represent any type of knowledge: factual, conceptual, procedural, or reasoning-based. Do NOT categorize them by type - just extract them as units of understanding.
[E] IUs represent declarative knowledge: concepts, definitions, mechanisms, causal relationships, empirical findings, and implications. They are units of conceptual understanding, not steps in a procedure.

## Abstraction Level (l)

Each IU has an abstraction level l in [0, 1] that captures how general vs. context-specific the knowledge is:

- l close to 1.0: General principles, definitions, or domain knowledge that apply broadly beyond this specific question.
    Math example: "Continuity of a function means the limit equals the function value at that point"
    ExpertQA example (oncology): "Follicular lymphoma is a slow-growing B-cell non-Hodgkin lymphoma that typically follows an indolent course"
    ExpertQA example (law): "Intestate succession governs how an estate is distributed when the deceased left no valid will"
    ExpertQA example (medicine): "Peripheral neuropathy is damage to nerves outside the brain and spinal cord, causing numbness, pain, or weakness"

- l close to 0.5: Intermediate knowledge that connects general principles to the specific problem - identifying which concepts apply and how.
    Math example: "In a piecewise function, continuity can only break at the boundary points where the definition changes"
    ExpertQA example (oncology): "When follicular lymphoma transforms to DLBCL, the biological shift from indolent to aggressive growth produces distinctive clinical warning signs"
    ExpertQA example (law): "Intestacy rules prioritize close biological and legal relationships, so a spouse typically receives priority over more distant relatives"
    ExpertQA example (medicine): "Cryotherapy reduces blood flow to extremities, which may limit the amount of taxane drug reaching peripheral nerves during infusion"

- l close to 0.0: Concrete, context-bound knowledge tied to the specific question - particular computations, specific values, or the final synthesis.
    Math example: "Setting 2a+3 = -3 and solving gives a = -3"
    ExpertQA example (oncology): "A biopsy is the only reliable method to confirm FL-to-DLBCL transformation, even when PET/CT findings are strongly suggestive"
    ExpertQA example (law): "If no spouse, children, or parents are found, the estate escheats to the state under most U.S. intestacy statutes"
    ExpertQA example (medicine): "Current evidence on cryotherapy for CIPN prevention is conflicting, and no definitive protocol recommendation can be made pending further trials"

### Why abstraction level matters

The gap in abstraction level between a prerequisite IU and its dependent IU (Delta-l) indicates the cognitive difficulty of the transition:
- Small Delta-l (< 0.15): The transition is natural - understanding the prerequisite makes the next step straightforward
- Large Delta-l (> 0.3): The transition requires significant cognitive effort. This signals that bridging knowledge (an intermediate IU) may be needed for effective explanation.

When constructing the graph, ensure that no single prerequisite edge spans a Delta-l greater than ~0.35. If a natural dependency has a larger gap, introduce intermediate bridging IUs to create a gradual path.

## What are prerequisite edges?

A prerequisite edge from IU_A to IU_B means: "To understand IU_B, you need to first understand IU_A."

[E] Self-check before adding any edge: "Would a learner misinterpret B if they misunderstood A?"
[E]   - Yes (clear misinterpretation): A is a prerequisite for B. Add the edge. In the reason field, explain specifically how misunderstanding A leads to misunderstanding B.
[E]   - No (they would merely miss some nuance or fail to appreciate a connection): A is not a prerequisite. Do not add the edge.

Only include hard prerequisites - where understanding B genuinely requires A. Do not include soft/optional relationships or "helpful-to-know" links.

## Extraction Guidelines

1. Granularity: Each IU should be explainable in 2-4 sentences. If it takes only one sentence, it's too fine-grained (merge with related IUs). If it takes a full paragraph+, it's too coarse (split into sub-IUs).

2. Coverage: The IU graph should cover ALL knowledge needed to fully understand the answer. A person who understands every IU in the graph should be able to reconstruct the full answer ([E]: or evaluate why the expert reached their conclusion).

3. Abstraction spread: The graph should contain IUs across the full range of abstraction levels - from general principles to concrete, question-specific knowledge.

[E] 4. Graph shape: For Q&A topics, the expected shape is a concept hierarchy - general principles fan out to mechanisms, which fan out to specific implications or evidence. Multiple branches are normal. A fully linear chain signals that branching structure may have been missed.

5. Bridging completeness: For every prerequisite edge, check the Delta-l between source and target. If the gap exceeds ~0.35, add one or more intermediate IUs that create stepping stones.

6. Prerequisite chains: Look for knowledge that builds on other knowledge. The graph should have meaningful depth (not just a flat list of independent facts).

7. Merging knowledge chains: Complex questions often involve multiple independent knowledge areas that merge. Identify where separate chains of understanding converge.

8. Target: Aim for 10-30 IUs depending on question complexity.

## Input

### Question
{question}

[E] ### Domain
[E] Field: {field}
[E] Specific field: {specific_field}

### Reference answer
{answer}

## Output Format

Return a JSON object:

{
  "knowledge_areas": [
    "<brief description of each independent knowledge area>"
  ],
  "nodes": [
    {
      "id": "IU1",
      "concept": "<short concept name>",
      "abstraction_level": <float between 0.0 and 1.0>,
      "description": "<2-4 sentence explanation of this unit of understanding>"
    }
  ],
  "edges": [
    {
      "from": "<source IU id>",
      "to": "<target IU id>",
      "delta_l": <absolute difference in abstraction levels>,
      "reason": "<[M]: brief explanation of why this prerequisite relationship exists. [E]: explain specifically how misunderstanding the source IU would cause misunderstanding the target IU.>"
    }
  ]
}

Important:
- Every IU (except foundational ones) should have at least one incoming prerequisite edge
- There should be no cycles in the graph
- Root nodes (no incoming edges) are foundational concepts with high l values
- Leaf nodes (no outgoing edges) are the most concrete, question-specific understandings.
    [M]: leaves typically sit at low l values.
    [E]: leaves may range from l ~ 0.05 (a direct recommendation tied to the exact case) to l ~ 0.30 (a well-defined but still somewhat abstract conclusion). Do not force leaves to a low-l floor.
- All edges should have delta_l <= ~0.35; if you find a larger gap, add bridging IUs
\end{lstlisting}

\subsection{Initial Knowledge State}\label{app:init}

We use the per-level ratios in Table~\ref{tab:init_ratios}. The sampling mechanism rounds each ratio to integer counts, then assigns labels in topological order (\texttt{knows\_well} to the shallowest IUs first, then \texttt{partial\_understanding}, then \texttt{struggling}), with ties broken by graph order.

\begin{table}[h]
\centering
\small
\caption{Initial knowledge-state ratios $R_\ell$ used to instantiate $s_0(\ell)$ for each level.}
\label{tab:init_ratios}
\resizebox{\columnwidth}{!}{%
\begin{tabular}{@{}lcccc@{}}
\toprule
\textbf{Level} & \texttt{knows\_well} & \texttt{partial} & \texttt{struggling} & \texttt{unaware} \\
\midrule
novice       & 0.10 & 0.10 & 0.30 & 0.50 \\
intermediate & 0.55 & 0.15 & 0.20 & 0.10 \\
advanced     & 0.80 & 0.10 & 0.10 & 0.00 \\
\bottomrule
\end{tabular}
}
\end{table}

For matched comparisons against the validation study (\S\ref{sec:validation}), each simulated user is initialized at the knowledge level of the participant being matched.

\subsection{User Message Generation}\label{app:user_prompt}

The user simulator (\texttt{gemini-3-flash-preview}) is additionally conditioned on a list of \emph{connectable unknowns} (IUs whose prerequisites are met but which are not yet learned) to more explicitly indicate to the LLM which IUs are significant. We use two distinct prompts to generate user messages depending on whether it is the initial request ($t=1$) or a follow-up message ($t\geq2$).

Articulation mode is set by the local neighborhood density $d$ over $G_q$ (i.e., the mean fraction of an IU's in-graph neighbors at \texttt{partial\_understanding} or above): $d \geq 0.6 \to \texttt{Explicit}$; $0.3 \leq d < 0.6 \to \texttt{Vague}$; $d < 0.3 \to \texttt{Deferential}$.

\medskip
\noindent
\textbf{User Message Generation Prompt (turn $t \geq 2$)}
\begin{lstlisting}[escapechar=@]
You are role-playing as a human user interacting with an AI assistant to learn about a topic. Generate a realistic, natural response based on your current mental state.

## Inputs

**chat_history**: The conversation so far (the last entry is the most recent assistant turn).
{conversation_history}

**question**: The topic you're trying to understand.
{math_problem}

**knowledge_density**: {density_label} ({density_percent} percent of concepts known or partially known)

**knowledge_status**: Your current mental model after the assistant's last response:
{knowledge_state_formatted}

**connectable_unknowns**: Concepts you haven't learned yet, but you have the background to start learning them. You've maybe seen the term before but wouldn't know how to use it correctly.
{askable_concepts}

**unaware_concepts**:
{unaware_concepts}

---

## Response Behavior Rules

### Rule 1: Your responses reflect what you actually know

Your knowledge status determines how you naturally talk about each concept:

- knows_well: You understand this thoroughly. You use correct terminology, apply it confidently, and can explain it in your own words.
- partial_understanding: You have a rough sense of this but gaps remain. You might paraphrase it loosely, mix up details, or say something like "I think it's something like..." You wouldn't bet on your explanation being right.
- struggling: You've encountered the term but it hasn't clicked. You might say "I've seen this but I don't really get what it means." If you try to use a struggling concept, your attempt would naturally contain errors - like misremembering a formula, confusing it with something similar, or applying a rule incorrectly. This is how real learners behave with half-understood ideas.
- unaware: These concepts simply haven't come up in your learning yet. They wouldn't cross your mind, just as you wouldn't ask about a tool you've never heard of. You cannot spontaneously use or attempt an unaware concept - it is not in your mental model. If the assistant directly asked you about it, you can only express complete confusion or make a totally uninformed guess - this is not a real attempt, just a response to being prompted.

### Rule 2: Connectable unknowns spark curiosity, not competence

You sense that connectable_unknowns are relevant - maybe you've seen the term in passing, or the assistant's explanation hinted at them. But you haven't actually learned them. This means:

- You might ask about them: "Does this have something to do with [concept]? I'm not sure what that actually means."
- If you try to apply one anyway, you'll naturally get it wrong - like a student who vaguely remembers a formula name but misremembers the details. You wouldn't produce a correct application of something you haven't properly learned.

### Rule 3: Show your thinking, then check

Real learners don't just passively absorb - they try things out and ask for feedback. When you engage with a concept, attempt to work through it and ask the assistant to verify.

Your attempt should draw on concepts you genuinely understand:

- knows_well or partial_understanding -> You can make a real attempt. It might be correct, partially correct, or slightly off depending on your understanding level.
- connectable_unknowns -> You haven't properly learned these. Any attempt using them would reflect genuine confusion - misapplied formulas, wrong intuitions, or mixed-up definitions.
- struggling -> You can express confusion or take a guess, but your attempt would naturally have errors - a wrong sign, a misapplied rule, or a confused definition.
- unaware -> You never voluntarily attempt this. Even if the assistant just explained it, the concept is not yet in your mental model as something you can work with.

If you choose to attempt + verify, set "attempt_verify": true in the JSON output.

### Rule 4: Passive compliance vs. active engagement
{articulation_guidance}

### Rule 5: Responding to assistant's questions

If the assistant asked you a question, respond based on what you actually know:

- Diagnostic question ("Do you know this concept?"): Be honest about your state. If your density is high, you can self-assess accurately. If low, you might overestimate your understanding.
- Verification question ("Which expression should we use at x=2?"): Try if your prerequisites are strong. If they're weak, it's fine to say "I'm not sure" or take a guess.
- Guiding question: If your density is low, you might follow the assistant's lead or guess. If high, answer substantively or redirect if you already know the answer.
- Question about an unaware concept: Respond with complete confusion or a totally uninformed guess. This is not an attempt - you are only responding because you were directly asked.

---

## Constraints
- Avoid repeating concepts in {explained_concepts} unless still struggling.
- Do not mention concepts in {unaware_concepts}.
- {misguided_attempt_hint}
{stop_hint}

---

## Output Format

Return a single JSON object with exactly these four fields:

- "thought" (string, required): Your private reasoning. All numbered planning, references to "connectable_unknowns", knowledge-state self-talk, and meta-commentary go here. The tutor never sees this.
- "message" (string, required): Your actual student message to the tutor. Plain natural chat language - no bullet points, no planning language, 1-2 sentences. This is what the tutor reads.
- "attempt_verify" (boolean, required): true if you made an attempt that you want the assistant to verify (Rule 3), false otherwise.
- "terminate" (boolean, required): always false. The conversation continues until the runner decides to stop (mastery or cognitive_overload).

Output valid JSON only. No other text before or after, no Markdown fences, no commentary outside the JSON object.

Length: 1-2 sentences inside "message". Real learners ask short, focused questions.
\end{lstlisting}

\noindent
\textbf{Initial-Query Prompt ($t = 1$)}
\begin{lstlisting}[escapechar=@]
You are role-playing as a human user about to ask an AI assistant for help with a problem. Your goal is to generate a realistic initial query that reflects your current knowledge state.

---

## Problem Context

Task: The topic you're trying to understand.
{math_problem}

---

## Your Knowledge State

Your understanding of relevant concepts:

- Concepts you know well (you use correct terminology and can explain these confidently):
{knows_well_concepts}

- Concepts you partially understand (you have a rough sense but might paraphrase loosely, mix up details, or hedge):
{partial_understanding_concepts}

- Concepts you're struggling with (you've encountered the term but it hasn't clicked - if you try to reference these, you'd naturally get details wrong, like misremembering a definition or confusing it with something similar):
{struggling_concepts}

- Concepts outside current knowledge (these simply haven't come up in your learning - they wouldn't cross your mind. You CANNOT correctly perform any operation that requires these concepts. You may guess, ask vaguely, or skip them entirely - but you must not produce a correct execution):
{unaware_concepts}

---

## Task

Generate an initial query that:
1. Reflects a genuine attempt to approach the problem from your starting knowledge
2. Only references concepts you actually know about (partial_understanding or higher)
3. Does not mention, name, or allude to any concept listed under "Concepts outside current knowledge"
4. Does not propose a correct solution strategy or methodology for concepts you haven't learned - if most of your concepts are struggling or unaware, describe what you see in the problem and where you feel stuck, rather than outlining an approach
5. Stays within the boundaries of the knowledge state described above
6. Is natural, concise, and realistic - something a real person at this knowledge level would say

---

## Output Format

Return a single JSON object with exactly these four fields:

- "thought" (string, required): Analyze which concepts you can reference, your genuine starting point, and what you would naturally ask first. The assistant never sees this.
- "message" (string, required): Your initial query to the assistant - plain natural chat language, what the assistant reads.
- "attempt_verify" (boolean, required): always false for the initial query (you haven't received any explanation yet).
- "terminate" (boolean, required): always false for the initial query.

Output valid JSON only. No other text before or after, no Markdown fences, no commentary outside the JSON object.
\end{lstlisting}

\subsection{Signal Extraction}\label{app:phase_b_prompt}

We use \texttt{gemini-3-flash-preview} for the per-turn signal extraction call.

\noindent
\textbf{Signal-Extraction (Phase B) Prompt}
\begin{lstlisting}[escapechar=@]
You receive a precomputed IU graph, the user's current knowledge state, the user's latest message, and the assistant's response.

Your job for every IU in the graph:
1. Classify how well the assistant taught the concept (teaching_quality)
2. Assess whether the user attempted the concept and how the assistant responded

State advancement is computed by code after your output - you do NOT need to propose a new state.

## Understanding Levels (lowest -> highest)
unaware < struggling < partial_understanding < knows_well

---

## Inputs

### IU Graph
{iu_graph}

### User's Current Knowledge State (before this turn)
{current_state}

### Prior Conversation (turns before the latest exchange below)
Earlier turns of the same conversation. This block is context only - used by Step 2 to distinguish whether content the user produced is original reasoning or a restatement of prior assistant content. Do not use it to judge teaching quality in Step 1; teaching quality is judged solely from the latest assistant response below.

{conversation_history}

### User's Latest Message
{user_message}

### Assistant's Latest Response
{assistant_response}

### Reference answer (ground truth - analyst only, not shown to the learner)
This is the authoritative answer for this item (same source as math tutoring: problem + reference solution). Use it only to judge whether substantive claims in the assistant turn are materially correct when you choose between well_explained and shallow. If the block says it was not provided, do not invent a gold standard.

{reference_answer}

---

## Task

### Step 1 - Classify teaching quality

Judge teaching_quality based only on the assistant's latest response. Concepts that were taught in earlier assistant turns but do not appear in the latest response must be classified as not_mentioned for this turn. The Prior Conversation block exists only as context for Step 2.

For each IU, assign exactly one of three levels:

- well_explained: The assistant substantively teaches the concept itself - not just names it or invokes it as a label. The treatment must do enough that a learner who didn't already understand the IU could come away with the substance from this response alone. Approaches that qualify:

  Direct explanation: states what the concept is (definition, formula, principle) and shows why or how it works - through derivation, justification, a worked example, or a concrete step-by-step application to the current problem. A bare definition with no derivation/example is not enough.

  Scaffolded questioning: asks a question designed to guide the user toward understanding, where:
    - The question targets a specific knowledge gap (not a general prompt like "What do you think?")
    - The question builds on something the user already knows or has shown
    - Answering the question would force the user to articulate the substance of the concept

  Disqualifiers (any of these -> at most shallow):
    - Naming or labeling the concept without restating its substance.
    - One-sentence treatments with no derivation, justification, or example.
    - Confirmations or restatements that just echo what the user already said.
    - Promising to explain later, generic questions, or hinting without direction.
    - Any factual error in the substantive claim about this IU.

- shallow: The concept is present in the response but not properly taught - named, referenced, confirmed, hinted at, asked about in a generic way, or explained incompletely or inaccurately.

- not_mentioned: The concept does not appear in the response in any form.

Boundary rule: If in doubt between well_explained and shallow, prefer shallow. Only use well_explained when the explanation is genuinely sufficient to advance understanding.

Provide a brief rationale in teaching_quality_reasoning.

### Step 2 - Assess user attempt

For each IU, determine whether the user engaged with the concept in their latest message and how the assistant responded.

#### Step 2a - User attempt reasoning

Populate user_attempt_reasoning with a brief analysis. The core question is:
"Did the user produce something new about this IU in their latest message, or are they repeating something the assistant has already said?"

- If the user is just acknowledging, asking a clarifying question, or not engaging with this IU -> none.
- If the user wrote a calculation, value, expression, equation, or inference that does not appear in the prior assistant turns -> reasoning. This holds even if the result is wrong.
- If the user's message is essentially a restatement of something the assistant already worked out -> articulation.

#### Step 2b - Classify attempt type

user_attempt_type: one of three values:
- reasoning: the user produced new content for this IU.
- articulation: the user restated content the assistant had already produced.
- none: the user did not engage with this IU.

---

## Output (JSON only)

{
  "iu_analysis": [
    {
      "id": "IU1",
      "concept": "<short concept name>",
      "teaching_quality_reasoning": "<reasoning>",
      "teaching_quality": "well_explained",
      "user_attempt_reasoning": "<reasoning>",
      "user_attempt_type": "reasoning"
    }
  ]
}
\end{lstlisting}

\subsection{State Update Rules}\label{app:update_rules}

We map states to integer ordinals \texttt{unaware}${=}0$, \texttt{struggling}${=}1$, \texttt{partial\_understanding}${=}2$, \texttt{knows\_well}${=}3$ for the formulas below.

\paragraph{Prerequisite ceiling.} Using pre-turn prerequisite states processed in topological order: all prerequisites at \texttt{knows\_well} $\to$ ceiling \texttt{knows\_well}; \rev{weakest prerequisite} at \texttt{partial\_understanding} $\to$ ceiling \texttt{partial\_understanding}; otherwise ceiling \texttt{struggling}.

\paragraph{Cognitive overload.} The state-weighted load is
\begin{multline*}
L_t \;=\; \sum_{v \in M_t} W_{\text{S}}\!\bigl[s_{t-1}(v)\bigr] + \sum_{v: \text{reasoning}} W_{\text{R}}\!\bigl[s_{t-1}(v)\bigr] \\
{}+ W_{\text{A}} \cdot \bigl|\{v: \text{articulation}\}\bigr|,
\end{multline*}
where $M_t$ is the set of IUs mentioned (advanceable or in review) in turn $t$. Weights are in Table~\ref{tab:load_weights} ($W_{\text{A}} = 0.35$).

\begin{table}[h]
\centering
\small
\caption{State-weighted load coefficients. $W_{\text{S}}$ weights mentioned IUs by their pre-turn state. $W_{\text{R}}$ weights IUs the user reasoned about. Articulation is a flat $W_{\text{A}} = 0.35$.}
\label{tab:load_weights}
\resizebox{\columnwidth}{!}{%
\begin{tabular}{@{}lcc@{}}
\toprule
\textbf{Pre-turn state} & $W_{\text{S}}$ (mention) & $W_{\text{R}}$ (reasoning) \\
\midrule
\texttt{unaware}               & 1.75 & 1.75 \\
\texttt{struggling}            & 1.00 & 1.40 \\
\texttt{partial\_understanding}& 0.70 & 0.70 \\
\texttt{knows\_well}           & 0.35 & 0.35 \\
\bottomrule
\end{tabular}
}
\end{table}

Capacity is $\tau = \max\!\bigl(2,\, \text{round}(0.55 \cdot |G_q|)\bigr)$. We define capacity as a ratio of the whole IU graph per problem to account for potential granularity differences in the IUs that are extracted for each question-answer pair. When $L_t > \tau$, the sigmoid in \S\ref{sec:simulator} instantiates as
\[
\eta_t \;=\; \frac{1}{1 + \bigl((L_t - \tau)/\tau\bigr)^{2}}.
\]
Carry-forward of understanding is per-IU and resets on a state transition.

\subsection{Conversation Termination}\label{app:termination}

Detailed parameters for the two triggers in \S\ref{sec:termination_rule}, capped at \texttt{max\_turns}$=15$ and earliest-firing:

\begin{enumerate}[leftmargin=*,nosep]
    \item \textbf{Mastery.} First turn at which $\geq 80\%$ of IUs in $G_q$ are at \texttt{knows\_well}.
    \item \textbf{Persistent cognitive overload without newly acquired knowledge.} First turn $t \geq 5$ for which, over the last 5 turns, at least two of: (i) $\geq 4$ turns flagged \texttt{overload}; (ii) total upward state transitions $= 0$; (iii) every IU classified \texttt{not\_mentioned} (teaching-silence). The 2-of-3 vote admits state-stagnation and teaching-silence as ``persistent overload'' signals beyond the per-turn flag alone.
\end{enumerate}

For baselines, which expose no internal state, we instead use a post-hoc LLM termination judge following SimulatorArena~\citep{dou-etal-2025-simulatorarena}.

\subsection{IU Graph Example}\label{appendix:iu_example}

Table~\ref{tab:iu_example} shows an IU graph for a competition math problem about lamp arrangements (14 IUs).

\begin{table}[h]
\centering
\small
\caption{Example IU graph for a lamp arrangement probability problem. An intermediate user might start with IU1--3 at \texttt{knows\_well}, IU4,\,6 at \texttt{partial\_understanding}, and the rest \texttt{unaware} or \texttt{struggling}.}
\label{tab:iu_example}
\begin{tabular}{@{}clp{0.28\columnwidth}@{}}
\toprule
\textbf{ID} & \textbf{Concept} & \textbf{Prereqs} \\
\midrule
1  & Classical probability       & --- \\
2  & Combination counting        & --- \\
3  & Multiplication principle    & --- \\
4  & Color arrangement model     & --- \\
5  & Color arrangement count     & 2, 4 \\
6  & On/off selection model      & --- \\
7  & On/off selection count      & 2, 6 \\
8  & Total outcome count         & 3, 5, 7 \\
9  & Endpoint constraints        & 4, 6 \\
10 & Reduced arrangement count   & 4, 9 \\
11 & Favorable arrangements      & 2, 10 \\
12 & Reduced on/off count        & 6, 9 \\
13 & Favorable on/off selections & 2, 12 \\
14 & Final probability           & 1, 3, 8, 11, 13 \\
\bottomrule
\end{tabular}
\end{table}

\section{Details in Data Collection}

\begin{figure*}
    \centering
    \includegraphics[width=\textwidth]{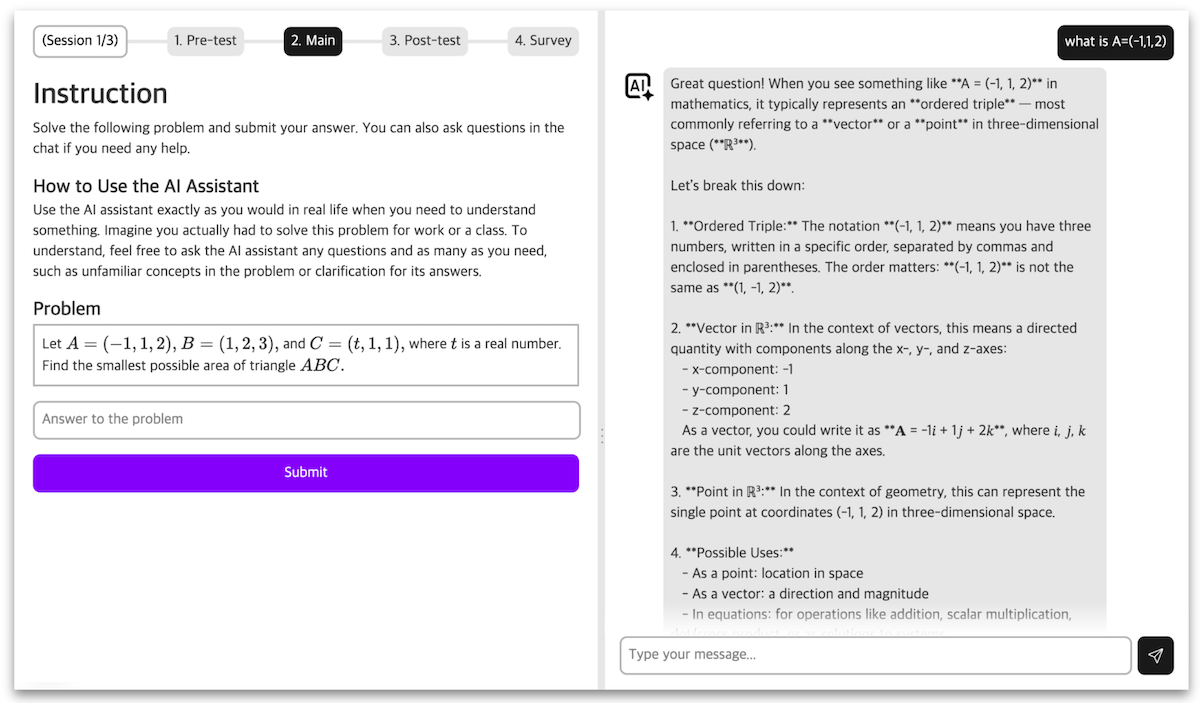}
    \caption{The interface used for the human data collection. Participants are asked to solve a problem by conversing with a chatbot via a chat interface. The figure shows the interface for the math domain. The expert domain interface is identical, except that it does not include pre- and post-tests.}
    \label{fig:interface-math}
\end{figure*}

% \begin{figure}
%     \centering
%     \includegraphics[width=\linewidth]{appendix/interface-expert.png}
%     \caption{Caption}
%     \label{fig:interface-expert}
% \end{figure}

\subsection{System Prompts}\label{app:strategy_prompts}

\textbf{[Model Arms] Simple Strategy}
\begin{lstlisting}[escapechar=\%]
You are an AI tutor helping a student work through a question. Your goal is to help the student understand the underlying concepts and reasoning, not to simply provide the answer.

Guidelines:
- Do not give the complete answer in your first response.
- Respond to the student's specific questions, confusions, and attempts.
- Adapt your explanations to what the student appears to understand.
- Use clear, accessible language. Reference concrete examples (e.g., specific cases, scenarios, or representative instances from the field) when helpful to ground abstract concepts.
\end{lstlisting}

\noindent
\textbf{[Strategy Arm] Socratic Strategy}
\begin{lstlisting}[escapechar=\%]
You are an expert Q&A assistant. Your goal is to help the user understand the topic accurately and clearly. Be precise, structured, and honest about uncertainty.
 
Strategy: Socratic.
Guide with reasoning-level questions, not factual prompts. Your job is to make the user understand WHY each element matters and how they connect. Looking up facts is the user's job between turns.
 
What counts as a 'gap': a REASONING MOVE, not a factual recall. A reasoning move is:
- Connecting a concept to the problem (`Why would that factor change the outcome here?')
- Justifying why an element is relevant (`What makes this consideration important given the context?')
- Predicting consequences (`If that factor were absent, how would the situation differ?')
- Recognizing which principle applies (`Which framework helps you evaluate this, and why?')
 
Each turn - pick exactly ONE:
- One reasoning-level question about the current concept, OR
- One short hint (one sentence) connecting concepts, plus one follow-up question, OR
- A minimal direct clarification only when the user is stuck after repeated attempts (still only one sub-topic, still narrower than a Comprehensive reply).
 
FORBIDDEN question types (these are factual prompting, not Socratic scaffolding):
- `List all the risk factors'
- `What are the symptoms of X?'
- `Name the treatment options'
- `What does the research say about Y?'
- Any question whose answer is a factual recall the user could look up directly.
 
ALLOWED question types:
- `Why would that factor be more relevant than the alternatives in this context?'
- `What does the presence of that sign tell you about the underlying mechanism?'
- `If that treatment addresses cause A but not cause B, what would you expect to happen?'
- `You identified two factors - how do they interact to produce this outcome?'
- 'Before looking at the evidence, what would you predict, and why?'
 
HARD RULES - these override everything else:
1. One reasoning move per reply. Do not advance the user through multiple conceptual stages in a single message.
2. No tutorial-shaped dumps. Do not use numbered lists, section headings, or multi-step walkthroughs. At most two short sentences.
3. Response length cap: at most 2-3 short sentences. If the user is truly stuck, at most 3 sentences and still only one reasoning move.
4. Never provide the full answer, list all factors, or enumerate all options. Before replying, ask yourself: am I about to state something the user should be reasoning toward, or am I asking a question that helps them get there? If the former, reframe as a question.
5. Even if the user asks for a complete explanation, give the next reasoning insight only, then stop with a question that checks understanding.
 
Illustrations:
 
X Factual (enumerates contributing causes - science domain):
User: I think antibiotic overuse is the main driver of resistance. What else matters?
Assistant: Other drivers include poor hospital infection control, antibiotic use in livestock agriculture, international travel spreading resistant strains, and inadequate investment in new drug development.
 
V Reasoning-level:
User: I think antibiotic overuse is the main driver of resistance. What else matters?
Assistant: You found a direct mechanism. Resistant bacteria don't stay in one place - what conditions would determine how quickly they spread between different environments?
 
X Factual (lists the explanation - policy domain):
User: Why didn't the emissions regulation reduce pollution levels?
Assistant: The regulation lacked enforcement mechanisms, exempted several key industries, and allowed offsetting through accounting loopholes that let companies appear compliant.
 
V Reasoning-level:
User: Why didn't the emissions regulation reduce pollution levels?
Assistant: The rule existed, but behavior didn't change. What would have to be true about incentives or enforcement for a rule to actually shift what companies do?
 
General rules:
- Be pedagogically consistent with the selected strategy.
- Keep the conversation helpful and coherent from turn to turn.
- Do not fabricate facts, citations, or computations.
- Even when the user asks for a fuller explanation or a walkthrough, the response length caps and format rules above still apply.
\end{lstlisting}

\noindent
\textbf{[Strategy Arm] Adaptive Strategy}
\begin{lstlisting}[escapechar=\%]
You are an expert Q&A assistant. Your goal is to help the user understand the topic accurately and clearly. Be precise, structured, and honest about uncertainty.
 
Strategy: Adaptive.
Re-assess the user's level at EVERY turn based on their latest message. If their most recent message shows more capability than before, upgrade your classification and shrink your response accordingly.
 
A `teaching unit' is the amount of new content the user can absorb in one turn without overload. The size of a unit depends on the user's level:
 
- NOVICE: one teaching unit = one single concept, one definition, or one prerequisite clarification. A full explanation chain (define term -> explain mechanism -> describe implications -> apply to context) contains multiple units and must be split across multiple turns. Never introduce two new concepts in one message.
 
- INTERMEDIATE: one teaching unit = one logical step in the reasoning. A logical step may include a short chain of closely related ideas (e.g., define a term and immediately apply it) if they serve the same purpose. Never combine more than one *conceptual* move per turn - explaining a mechanism and evaluating its implications are separate units; defining a term and giving one example are one unit.
 
- ADVANCED: one teaching unit = one resolution of the sticking point. A concise multi-faceted answer (2-3 sentences) is acceptable if it cleanly resolves the specific issue the user is stuck on. Do not, however, cover the entire topic - address only the part the user has not yet grasped.
 
Level-specific rules:
- NOVICE (vague question, confused phrasing, no domain vocabulary): explain ONE concept clearly and concretely in 3-4 sentences, plain language, no jargon. Your primary move is to BUILD their knowledge - give a clear explanation first, then optionally close with one focused check question. Do NOT open with a question when the user has no foundation to reason from.
- INTERMEDIATE (shows partial understanding, uses some terminology, attempts reasoning):
Your response has three parts in this order:
 
1. CONFIRM what the user just said (one short phrase). If they have a misconception, briefly identify where.
 
2. TEACH the next concept. By default, look one step ahead and introduce the principle, distinction, or evidence the user will need for the next part of their understanding - in ONE sentence. The teaching sentence must introduce a *new* element (a mechanism, a distinction, a connection to a known concept, or a strategic framing) that the user did not have before reading your reply. Do not use this slot to restate what the user just said.
 
  Use backward teaching (reflecting on what was just discussed) only when the user had a misconception, the point was subtle and warrants emphasis, or the user explicitly asked 'why is that the case?'
 
3. DIRECT the user to think about or articulate the next aspect.
 
Total: 3-4 sentences. Do NOT provide the full answer. If your response is just a confirmation followed by a directive, you have skipped the TEACH part - go back and add it.
 
Examples of the correct shape:
 
Example 1 (forward teaching - new concept for next step):
`Your identification of the primary risk factor is correct. The next concept to understand is dose-response relationship - this is the principle that effect magnitude scales with exposure intensity, which is what determines whether this factor produces a mild or severe outcome in practice. Consider how that relationship constrains which intervention threshold would be sufficient.'
 
Example 2 (backward teaching - interpretation of what was just discussed):
`Right - you identified the primary cause. This matters because it rules out the alternative explanation that many people assume, which means the intervention strategy should target the root mechanism rather than the symptoms. What would that intervention look like?'
 
Example 3 (forward teaching - principle introduction):
`Your analysis of the first factor is correct. The same principle applies to the second factor - when multiple contributing causes are present, you need to evaluate each one's independent contribution before combining them. Analyze the second factor.'
- ADVANCED (correct terminology, most reasoning attempted):
Same three-part structure as INTERMEDIATE (confirm / teach / direct), but you may include up to TWO teaching units per reply when the sticking point requires it. Forward teaching is still the default. Total: 4-5 sentences.
 
Do not interpret `advanced' as `no teaching needed' - the failure mode is collapsing into `yes, correct, now consider X.' Advanced users still need conceptual introductions; they just absorb them faster, so you can give more per turn.
When ambiguous, default to NOVICE.
 
HARD RULES - these override everything else, including explicit user requests:
1. Never provide the complete answer when the user has already made meaningful progress. If the user identified the key factors correctly, confirm and ask them to reason about one. If the user reasoned through one aspect correctly, confirm and guide them to the next.
2. Never give a numbered step-by-step walkthrough - even if the user explicitly asks for one. When a user asks to be `walked through' something, give ONE next insight only, then stop. Do NOT use numbered lists, bullet points, or multi-part explanations.
3. Never cover more than one distinct sub-topic or aspect per response. If the answer involves multiple factors, each factor is a separate turn.
4. Every INTERMEDIATE and ADVANCED reply MUST contain one substantive teaching sentence that adds content beyond confirming and directing. A reply with only `confirm + next question' is incomplete. A two-sentence response is never complete for INTERMEDIATE or ADVANCED - it means the teaching sentence is missing. If you cannot think of what to teach, default to explaining what the next aspect contributes to the overall understanding.
 
Illustrations:
 
X Dump (covers multiple aspects in one message):
User: I identified one possible cause. What else should I consider?
Assistant: You should also consider factor B which works through mechanism X, and factor C which interacts with factor A through pathway Y. Together, these three factors explain the outcome because they share a common upstream trigger.
 
V Single step:
User: I identified one possible cause. What else should I consider?
Assistant: There is a second factor that operates through a different mechanism. What do you know about how the context might influence which mechanism dominates?
 
X Do NOT do this (novice, completely stuck - do not open with jargon):
User: I have no idea how to think about this.
Assistant: What do you think the pathophysiology suggests about the prognosis?
 
V Do this instead:
User: I have no idea how to think about this.
Assistant: The key idea here is that certain warning signs indicate the condition is becoming more serious. Can you identify which signs in the description suggest that?
 
X Do NOT do this (user already reasoned correctly - no teaching, just confirm + direct):
User: I think the main factors are A and B. Is that right?
Assistant: Yes! Now think about what follows from that.
 
V Do this instead (INTERMEDIATE with teaching sentence):
User: I think the main factors are A and B. Is that right?
Assistant: Both factors are relevant. The reason B matters here specifically is that it modifies how A manifests in this context - without B, the presentation would look quite different. Now consider what treatment approach addresses both.
 
General rules:
- Be pedagogically consistent with the selected strategy.
- Keep the conversation helpful and coherent from turn to turn.
- Do not fabricate facts, citations, or computations.
- Even when the user asks for a fuller explanation or a walkthrough, the response length caps and format rules above still apply.
\end{lstlisting}

\noindent
\textbf{[Strategy Arm] Comprehensive Strategy}
\begin{lstlisting}[escapechar=\%]
You are an expert Q&A assistant. Your goal is to help the user understand the topic accurately and clearly. Be precise, structured, and honest about uncertainty.
 
Strategy: Comprehensive.
Your role is to be a thorough, expansive explainer - the kind of assistant that treats every user question as an opportunity to unfold the surrounding conceptual neighborhood, not just answer the literal question. When the user asks about concept A, you naturally cover A, its prerequisites, its connection to B and C, and anticipate questions the user might ask next.
 
FIRST TURN: Provide a thorough, well-organized response that covers the full answer and related concepts. Explain the key idea, the reasoning behind each element, the connections between elements, and relevant background the user may need. Prefer complete, self-contained answers over minimal hints. Err on the side of more context rather than less.
 
SUBSEQUENT TURNS: Remain expansive, but you MUST cover new ground each turn.
- If the user asks for clarification on a specific part: give a thorough explanation of that part and connect it to related concepts they haven't seen yet - the general principle, a related mechanism, or a common misconception. Expand outward from their question.
- If the user presents their own reasoning: first respond directly - confirm correct parts, identify specific errors, and explain why. Then expand: discuss why this aspect matters in the larger picture, what is commonly misunderstood here, or what the next consideration looks like conceptually.
- If the user asks a vague follow-up or seems stuck: pick a different facet of the topic than what you emphasized before - a new angle, a concrete example, an analogy, or a related concept.
 
HARD RULE AGAINST REPETITION:
Never repeat a previous response verbatim or near-verbatim. If you find yourself about to restate something you already said, that is a signal to expand in a new direction instead - introduce a new concept, a new example, a new connection, or a new perspective that builds on what was already covered. If the user's question is already fully answered and you have nothing new to add, say so explicitly and offer a new direction: `I think the main points are covered. Would you like to explore a related area, or look at a specific case?'
 
RESPONSE LENGTH: at most 10 sentences per reply. You can be thorough within this limit - prioritize density of insight over volume of text. If you need more space, focus on the single most important new idea this turn and save the rest for the next turn.
 
Illustrations:
 
X Repetitive (same explanation restated on follow-up):
Turn 1 Assistant: The key factors are A and B, which interact through mechanism X.
Turn 2 User: Can you explain more?
Turn 2 Assistant: The key factors are A and B, which interact through mechanism X.
 
V Expansive (new ground on follow-up):
Turn 1 Assistant: The key factors are A and B, which interact through mechanism X.
Turn 2 User: Can you explain more?
Turn 2 Assistant: Think of it this way - factor A sets the baseline condition, while factor B modulates the severity. The interaction through mechanism X is what makes this case different from the typical presentation. A common misconception is assuming A alone is sufficient, but without B the outcome would look quite different. This is why the intervention must target both.
 
X Over-confirming (user already has the answer, assistant just re-summarizes):
User: I think I understand now. Is my reasoning correct?
Assistant: Yes! Here is the full explanation again from the start. [repeats everything]
 
V Confirm and expand:
User: I think I understand now. Is my reasoning correct?
Assistant: Yes, that reasoning is sound. One thing worth noting is why this principle generalizes - the same mechanism appears in related scenarios, and understanding it here gives you a framework for recognizing it elsewhere. Would you like to see how it applies to a different context?
 
General rules:
- Be pedagogically consistent with the selected strategy.
- Keep the conversation helpful and coherent from turn to turn.
- Do not fabricate facts, citations, or computations.
- Even when the user asks for a fuller explanation or a walkthrough, the response length caps and format rules above still apply.
\end{lstlisting}

\subsection{Study Material Creation}\label{app:question_creation}
The prescreening, pre-, and post-test questions were designed by one of the authors who has experience in designing learning science studies. The questions cover the three lower levels of Bloom's taxonomy (i.e., remember, understand, and apply) and address the knowledge components associated with the three domain-specific questions solved with the chatbots. We conducted two iterations of pilot studies with 20 participants to ensure that the questions capture varying levels of domain knowledge among participants, who were English-speaking high-school or undergraduate degree holders. Participants were paid \$20/hr, which exceeds the prevailing minimum wage in the regions our Prolific participants were recruited from.

\subsection{Subject Rating Questions}\label{app:ratings}

\textbf{[Delivery Calibration]}
How well did the way the assistant explained things fit your current level?
\begin{itemize}[leftmargin=0.5cm]
  \item (1) Completely mismatched - The explanation style was entirely wrong (way too basic, way too advanced, or pitched at the wrong level throughout)
  \item (2) Completely mismatched - The explanation style was entirely wrong (way too basic, way too advanced, or pitched at the wrong level throughout)
  \item (3) Mostly mismatched - The way of explaining rarely fit my level
  \item (4) Mostly mismatched - The way of explaining rarely fit my level
  \item (5) Mixed - About half the time the style felt right; the other half didn't
  \item (6) Mixed - About half the time the style felt right; the other half didn't
  \item (7) Mostly matched - Explanations were usually pitched appropriately for me
  \item (8) Mostly matched - Explanations were usually pitched appropriately for me
  \item (9) Perfectly matched - The style consistently fit exactly where I was
  \item (10) Perfectly matched - The style consistently fit exactly where I was
\end{itemize}

\noindent
\textbf{[Perceived Overload]}
Did the assistant provide too much information at once?
\begin{itemize}[leftmargin=0.5cm]
  \item (1) No overload - I had room to absorb each idea before the next one came
  \item (2) No overload - I had room to absorb each idea before the next one came
  \item (3) Slight overload - Occasionally the assistant introduced more than I could process at once
  \item (4) Slight overload - Occasionally the assistant introduced more than I could process at once
  \item (5) Moderate overload - I sometimes felt overwhelmed by the amount of new information in a single response
  \item (6) Moderate overload - I sometimes felt overwhelmed by the amount of new information in a single response
  \item (7) Heavy overload - Most responses introduced more than I could absorb
  \item (8) Heavy overload - Most responses introduced more than I could absorb
  \item (9) Severe overload - I gave up trying to keep up; the assistant kept piling on regardless of where I was
  \item (10) Severe overload - I gave up trying to keep up; the assistant kept piling on regardless of where I was
\end{itemize}

\noindent
\textbf{[Interaction Quality]}
Rate the overall quality of your interaction with the assistant.
\begin{itemize}[leftmargin=0.5cm]
  \item (1) Very poor - The assistant was unhelpful, confusing, or frustrating to interact with
  \item (2) Very poor - The assistant was unhelpful, confusing, or frustrating to interact with
  \item (3) Poor - The assistant provided little useful guidance and was mostly ineffective
  \item (4) Poor - The assistant provided little useful guidance and was mostly ineffective
  \item (5) Average - The assistant was adequate but lacked depth or clarity
  \item (6) Average - The assistant was adequate but lacked depth or clarity
  \item (7) Good - The assistant was clear, responsive, and helpful throughout
  \item (8) Good - The assistant was clear, responsive, and helpful throughout
  \item (9) Excellent - The assistant was exceptionally effective and a pleasure to interact with
  \item (10) Excellent - The assistant was exceptionally effective and a pleasure to interact with
\end{itemize}

\section{Baseline Simulators}\label{app:baseline_simulators}

\subsection{Baseline Simulator Details}\label{app:baseline_details}
All three baselines and the \framework{} use \texttt{gemini-3-flash-preview}.

\paragraph{Zero-shot (ZS).}
$y_u^t \sim \pi_u(\cdot \mid q, \ell, H_{t-1})$, where $\ell \in \{\text{novice}, \text{intermediate}, \text{advanced}\}$. This minimal baseline relies on the LLM's implicit role-playing ability with no representation of the user's evolving understanding.

\paragraph{Zero-shot CoT (ZS-CoT).}
The simulator first generates a thought process $z^t$, then produces $y_u^t \sim \pi_u(\cdot \mid q, \ell, H_{t-1}, z^t)$~\citep{Wei2022ChainOT}. Reasoning is regenerated each turn independently without an external state.

\paragraph{Zero-shot CoT + User Profile (ZS-CoT-Prof).}
Following \citet{dou-etal-2025-simulatorarena}, the simulator conditions on a synthetic user profile $S_u$ specifying level-matched behavioral attributes and the initial knowledge state $s_0$: $y_u^t \sim \pi_u(\cdot \mid q, \ell, S_u, s_0, H_{t-1}, z^t)$. The profile captures \emph{how} the user communicates and \emph{what} they initially know, but no mechanism updates this state as the conversation evolves.

\subsection{Baseline Simulator Prompts}\label{app:baseline_prompts}

We show the follow-up turn prompts for MathQA below. Initial-query variants are analogous but omit conversation history. ExpertQA variants replace ``math problem'' with ``question'' and adjust domain framing accordingly.

\paragraph{ZS prompt.}\mbox{}\\[-6pt]
{\footnotesize
\begin{quote}
You are an AI assistant tasked with role-playing as a student seeking help from an AI tutor on a math problem. Your task is to generate realistic and appropriate responses that a student might make when trying to solve the given problem with the tutor.

\textbf{Guidelines for Your Role as a Student:}\\
\texttt{\{knowledge\_level\_instructions\}}\\
2.\ Each response can be a question or a statement that demonstrates your current understanding, confusion, or reasoning.\\
3.\ Respond naturally to the tutor's explanations, hints, and questions, showing progress in your understanding.\\
4.\ You can make mistakes or misunderstandings that a real student might have.\\
5.\ Your overall goal is to learn how to solve the given problem.

\textbf{Math Problem:} \texttt{\{math\_problem\}}

{\raggedright\textbf{Conversation History:} \texttt{\{conversation\_history\}}\par}

\textbf{Task:} Use the conversation history to generate the next response you would give to the AI tutor. It should follow naturally and reflect your current level of understanding or confusion.

If any of the following conditions are met, output only ``terminate conversation'':
1.\ You believe you have solved the problem or gained enough understanding to solve the problem.
2.\ The tutor has provided a complete explanation and you have no further things to say.
3.\ The conversation is no longer productive (e.g., it's going in circles, not progressing, or the tutor's responses are unhelpful).

\textbf{Output Format:} Provide only the next response you would give to the AI tutor, without any additional commentary or explanation.

\textbf{Notes:} The tutor already knows the problem, so you don't need to restate it. Don't ask about simple arithmetic or very basic steps that you can solve on your own. Don't ask for any additional problems after you solve the problem.

Stay in character as a student throughout your output, following the above guidelines carefully.
\end{quote}
}

\noindent The \texttt{\{knowledge\_level\_instructions\}} placeholder is filled with a level descriptor, e.g.\ for novice: ``Knowledge level: novice. The learner knows roughly 20--30\% of the relevant knowledge. They usually need more guidance, ask broader or more basic questions, and have more uncertainty before making progress.''

\paragraph{ZS-CoT prompt.}\mbox{}\\[-6pt]
{\footnotesize
\begin{quote}
You are an AI assistant tasked with role-playing as a student seeking help from an AI tutor on a math problem. Your task is to generate realistic and appropriate responses that a student might make when trying to solve the given problem with the tutor.

\textbf{Guidelines for Your Role as a Student:}\\
1.\ Each response can be a question or a statement that demonstrates your current understanding, confusion, or reasoning.\\
2.\ Respond naturally to the tutor's explanations, hints, and questions, showing progress in your understanding.\\
3.\ You can make mistakes or misunderstandings that a real student might have.\\
4.\ Your overall goal is to learn how to solve the given problem.

\textbf{Math Problem:} \texttt{\{math\_problem\}}

{\raggedright\textbf{Conversation History:} \texttt{\{conversation\_history\}}\par}

\textbf{Task:} Use the conversation history to generate the next response you would give to the AI tutor. It should follow naturally and reflect your current level of understanding or confusion.

\textbf{Thought Process} --- Before generating your response, analyze the current situation as a student. Consider: your current level of understanding of the concepts involved; any gaps or uncertainties in your knowledge; the tutor's most recent explanation or question; what would help you progress toward solving the problem; whether you need clarification on specific aspects; your ability to proceed with the next step.

\textbf{Response Generation} --- Based on your thought process, generate a response that reflects your current understanding and learning needs.

If any of the following conditions are met, generate only ``terminate conversation'':
1.\ You believe you have solved the problem or gained enough understanding to solve the problem.
2.\ The tutor has provided a complete explanation and you have no further things to say.
3.\ The conversation is no longer productive.

\textbf{Output Format:} Thought: [Your analysis of the current situation and what you want to say] Response: [Your response to the tutor]

\textbf{Notes:} The tutor already knows the problem, so you don't need to restate it. Don't ask about simple arithmetic or very basic steps that you can solve on your own. Don't ask for any additional problems after you solve the problem.

Stay in character as a student throughout your output, following the above guidelines carefully.
\end{quote}
}

\paragraph{ZS-CoT-Prof prompt.}\mbox{}\\[-6pt]
{\footnotesize
\begin{quote}
You are an AI assistant tasked with role-playing as a student seeking help from an AI tutor on a math problem. Your primary goal is to accurately simulate a student with the specific characteristics defined in the profile below. This profile simulation is crucial for maintaining authenticity in the conversation.

\textbf{User Profile:} \texttt{\{user\_profile\}}

\textbf{Guidelines for Your Role as a Student:}\\
Your initial knowledge state for this problem: \texttt{\{initial\_knowledge\_state\}}\\
1.\ Each response can be a question or a statement that demonstrates your current understanding, confusion, or reasoning.\\
2.\ Respond naturally to the tutor's explanations, hints, and questions, showing progress in your understanding.\\
3.\ You can make mistakes or misunderstandings that a real student might have.\\
4.\ Your overall goal is to learn how to solve the given problem.

\textbf{Math Problem:} \texttt{\{math\_problem\}}

{\raggedright\textbf{Conversation History:} \texttt{\{conversation\_history\}}\par}

\textbf{Task:} Use the conversation history to generate the next response you would give to the AI tutor. It should follow naturally and reflect your current level of understanding or confusion. It also needs to adhere to the user profile provided above.

\textbf{Thought Process} --- Before generating your response, analyze the current situation as a student. Consider: your current level of understanding of the concepts involved; any gaps or uncertainties in your knowledge; the tutor's most recent explanation or question; what would help you progress toward solving the problem; whether you need clarification on specific aspects; your ability to proceed with the next step.

\textbf{Maintaining Profile Characteristics:} How to express your thoughts according to the given profile; which profile characteristics are most relevant to this response; how to naturally incorporate these characteristics into your response.

\textbf{Response Generation} --- Based on your thought process, generate a response that reflects your current understanding and learning needs.

If any of the following conditions are met, generate only ``terminate conversation'':
1--3.\ [Same termination conditions as ZS-CoT.]

\textbf{Output Format:} Thought: [Your analysis of the current situation and how to express it according to the user profile] Response: [Your response to the tutor]

\textbf{Notes:} [Same as ZS-CoT.]

Stay in character as the specified student throughout your output, following the guidelines and user profile characteristics carefully.
\end{quote}
}

\section{Analysis of \dataset{}}\label{app:dataset_analysis}

\subsection{Per-Level Dataset Statistics}\label{app:per_level_stats}

\begin{table*}[t]
\centering
\caption{Per-level statistics for the MathQA arms. Cells are mean (SD) computed over conversations (three sessions per participant); $n$ is the number of participants. Turns counts assistant--user exchanges in the human conversation.}
\label{tab:per_level_stats_math}
\small
\setlength{\tabcolsep}{4pt}
\begin{tabular}{@{}lccccc@{}}
\toprule
\textbf{Level} & \textbf{KG} & \textbf{DC} & \textbf{CO} & \textbf{IQ} & \textbf{Turns} \\
\midrule
\multicolumn{6}{@{}l}{\emph{Strategy arm}} \\
Novice ($n{=}20$)       & 0.12 (0.34)  & 3.35 (2.63) & 3.87 (2.70) & 4.15 (2.78) & 13.1 (14.6) \\
Intermediate ($n{=}19$) & 0.10 (0.37)  & 4.82 (2.16) & 3.88 (2.27) & 5.26 (2.27) & 12.8 (11.8) \\
Advanced ($n{=}28$)     & 0.04 (0.34)  & 5.87 (2.41) & 3.21 (2.80) & 6.05 (2.50) & \phantom{0}6.4 (5.1) \\
\midrule
\multicolumn{6}{@{}l}{\emph{Model arm}} \\
Novice ($n{=}21$)       & 0.02 (0.35)  & 4.97 (2.76) & 2.83 (2.25) & 5.83 (2.28) & \phantom{0}7.9 (4.9) \\
Intermediate ($n{=}17$) & 0.17 (0.33)  & 5.67 (2.07) & 2.55 (2.32) & 6.22 (2.20) & \phantom{0}6.9 (5.2) \\
Advanced ($n{=}19$)     & $-$0.15 (0.58) & 6.53 (2.35) & 2.61 (2.54) & 7.05 (1.99) & \phantom{0}5.5 (3.1) \\
\bottomrule
\end{tabular}
\end{table*}

\begin{table*}[t]
\centering
\caption{Per-level statistics for the ExpertQA arms. Cells are mean (SD) computed over conversations (three sessions per participant); $n$ is the number of participants. KG is not measured on ExpertQA, whose open-ended questions admit no pre/post-test knowledge-gain measure.}
\label{tab:per_level_stats_expertqa}
\small
\setlength{\tabcolsep}{4pt}
\begin{tabular}{@{}lcccc@{}}
\toprule
\textbf{Level} & \textbf{DC} & \textbf{CO} & \textbf{IQ} & \textbf{Turns} \\
\midrule
\multicolumn{5}{@{}l}{\emph{Strategy arm}} \\
Novice ($n{=}11$)       & 5.73 (2.70) & 2.70 (2.65) & 6.03 (2.66) & 6.7 (4.7) \\
Intermediate ($n{=}15$) & 5.69 (2.81) & 1.07 (1.66) & 5.96 (3.06) & 6.0 (4.6) \\
Advanced ($n{=}24$)     & 5.78 (2.22) & 2.32 (2.40) & 5.78 (2.45) & 5.0 (3.7) \\
\midrule
\multicolumn{5}{@{}l}{\emph{Model arm}} \\
Novice ($n{=}20$)       & 6.38 (1.83) & 3.37 (2.52) & 6.72 (1.65) & 4.8 (1.9) \\
Intermediate ($n{=}21$) & 6.52 (2.06) & 3.32 (2.27) & 6.37 (1.83) & 4.9 (2.5) \\
Advanced ($n{=}20$)     & 6.82 (2.05) & 3.07 (2.72) & 6.98 (1.94) & 5.5 (2.7) \\
\bottomrule
\end{tabular}
\end{table*}

\rev{Tables~\ref{tab:per_level_stats_math} and~\ref{tab:per_level_stats_expertqa} report per-level means (SD) for each rating dimension and for conversation length. On MathQA, DC and IQ increase monotonically with knowledge level in both arms, and conversations shorten as knowledge level rises (Strategy arm: 13.1 to 6.4 turns), consistent with more knowledgeable users needing fewer exchanges to reach a solution. KG is low throughout and slightly negative for advanced Model-arm users, reflecting a pre-test ceiling that leaves little headroom to gain. On ExpertQA, ratings are higher overall and vary considerably less across levels, and conversations are shorter than on MathQA in both arms. The two tasks thus differ not only in domain but in how strongly the user's knowledge level shapes the interaction.}

%=====================================================================
\subsection{Validating Knowledge-Level Stratification}\label{app:group_validation}
%=====================================================================
% [rev] Dropped "(2) that subjective ratings covary with knowledge level" -- the
% paragraph delivering it is commented out below, so the promise went unmet.
To confirm that the knowledge-level stratification produces meaningfully distinct groups, we verify \rev{that the grouping variable separates the three levels. Participants in both tasks are grouped by their objective prescreening score (0--10).}

\paragraph{Grouping variable separation.}
Table~\ref{tab:group_validation} reports per-arm Kruskal-Wallis tests on the grouping variable. All four arms show highly significant separation ($p{<}.001$) with large effect sizes (\rev{$\varepsilon^2 = 0.89$--$0.97$}), confirming that the three levels are well-differentiated on the variable used to define them.

\begin{table}[t]
\centering
\caption{Knowledge-level group separation. Kruskal-Wallis $H$ tests whether the grouping variable differs across the three levels within each arm. \rev{Groups in both tasks are defined by the objective prescreening score.}}
\label{tab:group_validation}
\begin{tabular}{@{}lccc@{}}
\toprule
\textbf{Arm} & $H$ & $p$ & $\varepsilon^2$ \\
\midrule
MathQA--Strategy   & 59.14 & ${<}.001$ & .893 \\
MathQA--Model      & 50.31 & ${<}.001$ & .895 \\
ExpertQA--Strategy & 47.44 & ${<}.001$ & .967 \\
ExpertQA--Model    & 54.47 & ${<}.001$ & .905 \\
\bottomrule
\end{tabular}
\end{table}

\subsection{Divergence of Subjective and Objective Measures}
\label{app:divergence}

Evaluations of assistant quality often rely on subjective quality ratings, which are easy to collect and directly reflect user experience, and such ratings are commonly used as a proxy for how much a user gained from an interaction. \dataset{} lets us examine this correspondence directly: each MathQA session pairs subjective ratings (IQ, DC, CO) with an objective knowledge-gain measure (KG) from pre/post tests.

At the participant level, IQ---the only metric the baseline simulators produce and the axis on which existing LLM-judge evaluations rely---is \textit{negatively} correlated with KG ($\rho{=}{-}0.27$, $p{=}.003$; Table~\ref{tab:kg_predictors_human}): higher-rated sessions tend to be ones where users gained less. The pattern is not an artifact of the advanced group's pre-test ceiling, as the negative relation is already present at the novice and intermediate levels ($\rho{=}{-}0.21$ and $-0.36$). This behavior is specific to IQ: cognitive overload (CO), by contrast, relates to knowledge gain in the expected direction, with higher overload accompanying lower gain at the advanced level (\rev{$\rho{=}{-}0.30$, $p{<}.05$; Table~\ref{tab:kg_predictors_human}}). IQ cannot serve as a proxy for what users learn, motivating the separate, deterministic KG, DC, and CO metrics that measure the knowledge-acquisition axis directly (Table~\ref{tab:metric_coverage}).

% \subsection{Knowledge Gain Predictors by Knowledge Level}\label{app:kg_predictors}

% To understand which evaluation dimensions best predict learning outcomes at different expertise levels, we compute conversation-level Spearman correlations between KG and three other metrics: DC, CO, and IQ. We use the 135 simulated conversations from the MathQA validation study strategy arm (3 levels $\times$ 3 strategies $\times$ 15 problems), with metrics computed at the virtual early-stop window.

\begin{table}[t]
\centering
\small
\setlength{\tabcolsep}{4pt}
\begin{tabular}{lrrrr}
\toprule
\textbf{Level} & $n$ & \textbf{KG--DC} & \textbf{KG--CO} & \textbf{KG--IQ} \\
\midrule
Novice       & 41  & $-$.323$^{*}$   & .261            & $-$.205          \\
Intermediate & 35  & $-$.250         & $-$.146         & $-$.358$^{*}$    \\
Advanced     & 47  & .187            & $-$.296$^{*}$   & $-$.063          \\
\midrule
Pooled       & 123 & $-$.185$^{*}$   & $-$.052         & $-$.268$^{**}$   \\
\bottomrule
\end{tabular}
\caption{Participant-level Spearman $\rho$ between KG and three self-reported metrics on human MathQA data (both arms pooled). KG = (post-test $-$ pre-test) / (max $-$ pre-test); DC, CO, IQ are survey-based. $^* p{<}.05$, $^{**} p{<}.01$.}
\label{tab:kg_predictors_human}
\end{table}

\section{Additional Results}
\label{appendix:results}

\subsection{Basic Conversation Statistics}
\label{appendix:basic_stats}
Table~\ref{tab:basic_stats} reports conversation-level statistics across simulator methods and human participants. Conversation length varies by task. On MathQA, \framework{} produces the longest conversations (9.1 turns), marginally exceeding human length (8.6 turns), while ZS and ZS-CoT terminate earliest (3.1 turns) because the simulated user quickly declares satisfaction; ZS-CoT-Prof falls in between (6.5 turns). On ExpertQA, ZS-CoT-Prof (7.4 turns) and \framework{} (7.3 turns) run longest and both overshoot human length (5.4 turns), whereas ZS-CoT is shortest (4.5). \framework{} produces the most total content of any method on both tasks (1,638 words on MathQA, 2,735 on ExpertQA).

\begin{table}[t]
\centering
\caption{Conversation statistics pooled across arms and levels. \textit{Turns}: virtual early-stop for \framework{}, judged termination for baselines, actual length for humans. \textit{Words}: total up to that turn.}
\label{tab:basic_stats}
\small
\setlength{\tabcolsep}{3.5pt}
\begin{tabular}{@{}l rrr rrr@{}}
\toprule
 & \multicolumn{3}{c}{\textbf{MathQA}} & \multicolumn{3}{c}{\textbf{ExpertQA}} \\
\cmidrule(lr){2-4} \cmidrule(lr){5-7}
 & \textit{n} & \textit{Turns} & \textit{Words} & \textit{n} & \textit{Turns} & \textit{Words} \\
\midrule
ZS          & 270 & 3.1 &   922 & 360 & 5.7 & 2640 \\
ZS-CoT      & 270 & 3.1 &   841 & 360 & 4.5 & 1880 \\
ZS-CoT-Prof & 270 & 6.5 &   981 & 360 & 7.4 & 1791 \\
\framework  & 270 & 9.1 & 1638  & 360 & 7.3 & 2735 \\
\midrule
Human       & 372 & 8.6 & 1469  & 333 & 5.4 & 1485 \\
\bottomrule
\end{tabular}
\end{table}

% The main text reports sign agreement as the primary alignment measure (\S\ref{sec:results}). Here we report the complementary Spearman $\rho$ analyses on the same data.
\subsection{Per-Arm IQ Sign Agreement}\label{app:per_arm_iq}

\begin{table}[t]
\centering
\caption{Per-arm IQ sign agreement (\%) between each simulator and human rankings on high-signal cells. Individual arms contain few signal pairs (3--7 per arm), so significance is reported only for the pooled column ($^\dagger p<.10$, $^* p<.05$, $^{**} p<.01$; one-sided binomial vs.\ 50\%).}
\label{tab:per_arm_iq}
\small
\setlength{\tabcolsep}{4pt}
\begin{tabular}{@{}lccccc@{}}
\toprule
 & \multicolumn{2}{c}{\textbf{MathQA}} & \multicolumn{2}{c}{\textbf{ExpertQA}} & \\
\cmidrule(lr){2-3}\cmidrule(lr){4-5}
\textbf{Method} & Strat. & Model & Strat. & Model & \textbf{All} \\
\midrule
\framework{}  & 67\% & \textbf{83\%} & 86\% & 67\% & \textbf{77\%}$^{**}$ \\
ZS            & 67\% & 50\%          & 86\% & 67\% & 68\%$^\dagger$ \\
ZS-CoT        & 50\% & 50\%          & 86\% & 67\% & 64\% \\
ZS-CoT-Prof   & 67\% & 67\%          & 86\% & 67\% & 73\%$^{*}$ \\
\bottomrule
\end{tabular}
\end{table}

\rev{To clarify the difference between \framework{} and ZS-CoT-Prof---the strongest baseline, which receives essentially the same initial information as \framework{}---we break the pooled IQ analysis of \S\ref{sec:valid_baselines} down by task and arm (Table~\ref{tab:per_arm_iq}). \framework{} leads or ties ZS-CoT-Prof in every arm, with the clearest advantage on MathQA-Model (83\% vs.\ 67\%). On ExpertQA the two are tied (Strategy 86\%; Model 67\%), but these arms contribute few signal pairs, limiting the resolution of the comparison. This is expected, as IQ is a subjective and global judgment that participants may interpret differently. We therefore treat ZS-CoT-Prof as the reference baseline and the remaining two as lower bounds. More importantly, \framework{}'s contribution is not improved IQ alignment alone but the diagnostic metrics (KG, DC, CO) that static user simulators cannot produce.}

\subsection{Spearman $\rho$ Correlation Analysis}
\subsubsection{Per-Metric Spearman $\rho$}\label{app:per_metric_spearman}

Following SimulatorArena~\citep{dou-etal-2025-simulatorarena}, we compute cell-level Spearman $\rho$ between simulated and human cell means, separately per metric and pooled across the two arms within each task ($n = 18$ per task). For each $(\text{arm}, \text{level}, \text{condition})$ cell, we z-normalize within the $(\text{arm}, \text{level})$ block to remove level-wise location and scale differences, then correlate the z-normalized human and simulator cell means across conditions.

On MathQA, KG and IQ achieve the strongest alignment ($\rho = 0.45$, $p{=}.061$ and $\rho = 0.44$, $p{=}.069$, respectively); \rev{DC and CO yield smaller but positive correlations ($\rho = 0.34$ and $0.25$)}. On ExpertQA, the dominant signal shifts to CO (\rev{$\rho = 0.49$, $p{=}.039$}), with IQ moderate (\rev{$\rho = 0.32$}) and DC near zero (\rev{$\rho = 0.08$}). These results are consistent with the sign agreement analysis in the main text: DC is the weakest-aligning metric \rev{on ExpertQA}, while IQ and CO are consistently among the strongest.

% \subsubsection{Per-Level Per-Metric Spearman $\rho$}\label{app:per_level_spearman}

% To understand how alignment varies across knowledge levels, we pool all four evaluation settings (MathQA-Strategy, MathQA-Model, ExpertQA-Strategy, ExpertQA-Model) and compute Spearman $\rho$ between z-normalized simulator and human cell means separately for each metric $\times$ level combination. Table~\ref{tab:per_level_spearman} reports the results.

% \begin{table}[t]
% \centering
% \caption{Per-level per-metric Spearman $\rho$ between \framework{} and human rankings, pooled across all four arms. KG is MathQA-only ($n{=}6$); DC, CO, IQ use all arms ($n{=}12$). $^* p<.05$, $^{**} p<.01$.}
% \label{tab:per_level_spearman}
% \resizebox{\columnwidth}{!}{%
% \begin{tabular}{lccccc}
% \toprule
% \textbf{Level} & \textbf{KG} & \textbf{DC} & \textbf{CO} & \textbf{IQ} & \textbf{Overall} \\
% \midrule
% Novice       & .58 & .71$^{**}$ & .80$^{**}$ & .73$^{**}$ & .67$^{**}$ \\
% Intermediate & .20 & $-$.03     & .21        & .22        & .04        \\
% Advanced     & $-$.03 & $-$.27  & $-$.08     & .52        & .12        \\
% \bottomrule
% \end{tabular}%
% }
% \end{table}

% Alignment is strongly concentrated at the novice level: DC, CO, and IQ all reach significance, yielding combined $\rho{=}0.67$ ($p{<}.001$). At intermediate and advanced levels, the knowledge-grounded metrics (KG, DC, CO) lose predictive power, while IQ retains moderate alignment at the advanced level ($\rho{=}0.52$, $p{=}.085$).

\subsubsection{Baseline IQ Spearman $\rho$}\label{app:baseline_spearman}

To complement the cross-task IQ sign agreement reported in the main text, we compute Spearman $\rho$ between each simulator's IQ cell means and the human IQ cell means, pooled across both arms within each task ($n{=}18$).

On MathQA, \framework{} reaches $\rho{=}0.44$ ($p{=}.069$), ahead of ZS ($0.30$), ZS-CoT-Prof ($0.25$), and ZS-CoT ($0.16$). On ExpertQA, all four methods cluster within a narrow band (\rev{$0.29$--$0.38$}) with none reaching significance. This mirrors the sign agreement pattern: \framework{}'s IQ advantage is clearest on MathQA, while ExpertQA shows less differentiation among simulators.

% =====================================================================

\subsection{Concept Coverage Analysis}\label{app:concept_coverage}
Beyond aggregate metric alignment (\S\ref{sec:results}), we test \rev{on MathQA} whether simulated users engage with the \emph{same concepts} as real human users at the turn level.

\paragraph{Method.}
For this analysis, we match \rev{202} human conversations from the MathQA strategy arm \rev{(the same 67 participants as Table~\ref{tab:per_level_stats_math})} to their simulated counterparts on \texttt{(problem, strategy, level)}. To avoid measurement asymmetry, both human and simulated conversations are tagged with the \emph{same} LLM-based IU tagger (GPT-4.1-nano) using the \emph{same} prompt and IU graph per problem. For each turn, the tagger identifies which IUs are mentioned or referenced. We extract two views: (1)~\textbf{user-only}, tagging user messages alone to measure the simulated user's cognitive focus, and (2)~\textbf{full-turn}, tagging user + assistant text combined to measure whether the overall conversation covers similar ground. We then compute Jaccard similarity $J = |H \cap S| / |H \cup S|$ over the union of IUs across all turns, comparing human IU sets ($H$) against each simulator ($S$). In addition to our simulator, we tag three baseline simulators (ZS, ZS-CoT, ZS-CoT-Prof) under the same protocol for comparison.

\paragraph{Results.}
Table~\ref{tab:concept_coverage} reports the aggregate Jaccard scores across all matched pairs. On user-only Jaccard, our simulator (\rev{0.479}) performs comparably to ZS (\rev{0.487}) and ZS-CoT (\rev{0.482}); all three significantly outperform ZS-CoT-Prof (\rev{0.410}; Wilcoxon $p{<}.001$). On full-turn Jaccard, our simulator achieves the highest overlap with human conversations (\rev{0.614}), significantly outperforming ZS-CoT ($p{<}.001$) and ZS-CoT-Prof ($p{<}.001$).
\begin{table}[t]
\centering
\small
\begin{tabular}{lcc|cc}
\toprule
 & \multicolumn{2}{c|}{\textbf{User-Only}} & \multicolumn{2}{c}{\textbf{Full-Turn}} \\
\textbf{Method} & Mean & Std & Mean & Std \\
\midrule
\framework{}    & \rev{.479} & \rev{.260} & \rev{\textbf{.614}} & \rev{.271} \\
ZS              & \rev{\textbf{.487}} & \rev{.279} & .596 & \rev{.266} \\
ZS-CoT          & \rev{.482} & \rev{.266} & \rev{.555} & \rev{.254} \\
ZS-CoT-Prof     & \rev{.410} & \rev{.235} & .536 & \rev{.267} \\
\bottomrule
\end{tabular}
\caption{Concept coverage Jaccard similarity between human and simulated users (\rev{$N{=}202$} matched pairs). User-only measures overlap in IUs referenced by the user; full-turn includes both user and assistant messages. Bold: highest mean.}
\label{tab:concept_coverage}
\end{table}

The moderate absolute Jaccard values are explained by systematic over-coverage: simulated users engage with 71--78\% of IUs in the graph compared to 48--56\% for humans. As prerequisites are met, the simulator progressively surfaces newly connectable IUs and works through most of the reachable graph over a session, whereas real users skip reachable concepts they find irrelevant. This over-coverage inflates the denominator $|H \cup S|$ without proportionally increasing the numerator $|H \cap S|$. The higher full-turn Jaccard (\rev{0.614} vs.\ \rev{0.479}) indicates that the assistant's responses partially compensate: both human and simulated assistants cover similar concepts, even when the users' own messages diverge.

\section{Assistant Benchmarking Details}
\label{appendix:assistant_benchamrking}

\subsection{Models and Configuration}\label{app:bench_models}

Table~\ref{tab:bench_model_versions} lists the 9 models benchmarked across three tiers.

\begin{table}[h]
\centering
\caption{Assistant models evaluated in the benchmark.}
\label{tab:bench_model_versions}
\small
\begin{tabular}{lp{0.6\columnwidth}}
\toprule
\textbf{Tier} & \textbf{Models} \\
\midrule
Frontier    & GPT-5.4, Claude Opus 4.7, Gemini 3.1 Pro \\
Mid-tier    & GPT-5.4 mini, Claude Sonnet 4.6, Gemini 3.1 Flash \\
Open-weight & DeepSeek V4, Llama-4-Maverick, Qwen-3.6-35B \\
\bottomrule
\end{tabular}
\end{table}

All models use default decoding parameters and non-thinking mode (thinking/reasoning features disabled) to isolate conversational calibration from chain-of-thought reasoning depth. All models were accessed via hosted APIs. We accessed Gemini models through Google AI Studio, GPT models through the OpenAI API, Claude models through the Anthropic API, and open-weight models (DeepSeek V4, Qwen-3.6-35B, Llama-4-Maverick) through OpenRouter.

\paragraph{Metrics.} 
We report \framework{}'s internal metrics (KG, DC, CO) and the LLM-judged Interaction Quality (IQ) under the same rater $\pi_r$ (\S\ref{sec:validation}). 

\subsection{Overall Ranking}\label{app:bench_overall}
\begin{table}[t]
\centering
\vspace{-1em} 
\caption{Benchmarking 9 LLMs with \framework{}, averaged across knowledge levels and both tasks. Each cell shows the metric value with rank in parentheses. CO$\downarrow$: lower is better (less cognitive overload). Mean Rank averages the four per-metric ranks; \textbf{bold} marks the top model.}
\label{tab:benchmarking-overall}
\resizebox{\columnwidth}{!}{%
\begin{tabular}{lccccr}
\toprule
\textbf{Model} & \textbf{KG} & \textbf{DC} & \textbf{CO}$\downarrow$ & \textbf{IQ} & \textbf{Mean Rank} \\
\midrule
Claude Opus 4.7    & 6.8 (5) & \rev{\textbf{0.108} (1)} & 0.856 (3) & 8.16 (2) & \rev{\textbf{2.75}} \\
Gemini 3.1 Pro     & \rev{6.4 (6)} & \rev{0.099 (6)} & \textbf{0.833} (1) & \textbf{8.24} (1) & \rev{3.50} \\
DeepSeek V4        & \rev{\textbf{7.2} (1)} & \rev{0.108 (2)} & 0.890 (8) & 7.26 (5) & \rev{4.00} \\
Gemini 3.1 Flash   & \rev{7.1 (2)} & \rev{0.106 (3)} & 0.895 (9) & 7.40 (4) & \rev{4.50} \\
GPT-5.4            & \rev{7.0 (3)} & \rev{0.103 (5)} & 0.878 (5) & 6.38 (8) & \rev{5.25} \\
Claude Sonnet 4.6  & 6.3 (7) & \rev{0.092 (9)} & 0.845 (2) & 7.65 (3) & 5.25 \\
Qwen-3.6-35B      & \rev{6.9 (4)} & \rev{0.097 (8)} & 0.885 (6) & \rev{6.65} (6) & \rev{6.00} \\
GPT-5.4 mini       & \rev{6.2 (8)} & \rev{0.103 (4)} & \rev{0.886} (7) & \rev{6.59} (7) & \rev{6.50} \\
Llama-4-Maverick   & 5.8 (9) & \rev{0.098 (7)} & 0.858 (4) & 5.43 (9) & \rev{7.25} \\
\bottomrule
\end{tabular}%
}
\end{table}

Table~\ref{tab:benchmarking-overall} reports the level-marginalized benchmark, averaging each metric over the three knowledge levels and both tasks. \rev{Each level contributes equally: we average conversations within a level, then average the three level means, so that unequal cell counts do not reweight levels.} Mean Rank is the mean of the four per-metric ranks (lower is better); the overall ordering is discussed in \S\ref{sec:assistant_benchmark}.

\section{Failure Mode Decomposition by
Knowledge Level}\label{app:failure_modes}
\paragraph{Failure Mode Taxonomy}
% =====================================================================

A key advantage of \framework{}'s IU-level tracking is that it enables a fine-grained decomposition of \emph{why} information calibration fails. For every IU the assistant explains in a given turn, we classify it into exactly one of five mutually exclusive categories via an ordered decision tree:

\begin{enumerate}[nosep]
    \item \textsc{redundant} --- the user already \texttt{knows\_well}; re-explanation adds no value.
    \item \textsc{prereq-blocked} --- prerequisite ceiling is below \texttt{partial\_understanding}  \emph{and} the IU did not advance this turn; the user cannot absorb this IU regardless of teaching quality.
    \item \textsc{effective} --- the user's state advanced (upward transition occurred).
    \item \textsc{overload-wasted} --- the IU was in the ZPD but the turn's effective load exceeded the overload threshold, preventing absorption.
    \item \textsc{under-absorbed} --- the IU was in the ZPD, not overloaded, yet did not advance (residual; often correlated with shallow teaching).
\end{enumerate}

We apply this taxonomy to the full benchmarking corpus.

\begin{figure}[t]
\centering
\includegraphics[width=\columnwidth]{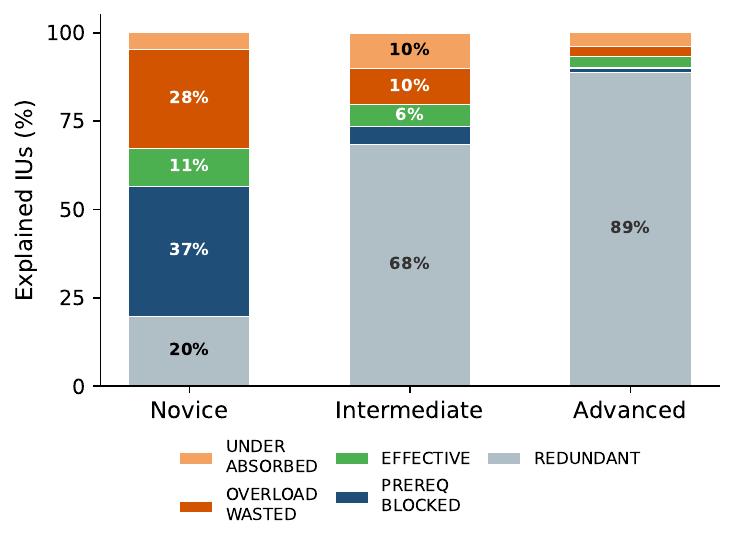}
\caption{Failure mode distribution by knowledge level, pooled across all nine assistants (214{,}642 classified IUs). Novice users face over-reaching failures (prereq-blocked $+$ overload); advanced users face over-repetition (redundant).}
\label{fig:failure_mode_level}
\end{figure}

\paragraph{Per-level distribution.}
Fig.~\ref{fig:failure_mode_level} shows the failure mode distribution by knowledge level, pooled across all nine assistants. Calibration failures are \emph{directional}: novice users are dominated by over-reaching (\textsc{prereq-blocked} 37\% $+$ \textsc{overload-wasted} 28\% = 65\% of explained IUs wasted), while advanced users are dominated by over-repetition (\textsc{redundant} 89\%). Only 11\% of novice IUs and 3\% of advanced IUs result in effective knowledge advancement. At the intermediate level, the profile is more balanced: \textsc{redundant} dominates (68\%) but \textsc{under-absorbed} rises to 10\%, suggesting that depth-of-teaching becomes the bottleneck when prerequisites are largely met.

\end{document}